\documentclass[10pt,twocolumn,letterpaper]{article}

\usepackage{cvpr}              
\usepackage{multirow}
\usepackage{pifont}
\usepackage{algorithmic}
\usepackage{algorithm}
\usepackage[labelfont=bf]{caption}
\usepackage[table]{xcolor}

\definecolor{cvprblue}{rgb}{0.21,0.49,0.74}
\usepackage[pagebackref,breaklinks,colorlinks,allcolors=cvprblue]{hyperref}

\def\paperID{***} 
\def\confName{CVPR}
\def\confYear{2026}

\title{Refining Few-Step Text-to-Multiview Diffusion via Reinforcement Learning}

\author{
Ziyi Zhang$^{1}$ \quad
Li Shen$^{2}$ \quad
Deheng Ye$^{3}$ \quad
Yong Luo$^{1}$\footnotemark[1] \\
Huangxuan Zhao$^{1}$\footnotemark[1] \quad
Meng Liu$^{4}$ \quad
Wei Yu$^{1}$\footnotemark[1] \quad
Lefei Zhang$^{1}$ \\[0.5em]
$^{1}$ School of Computer Science, National Engineering Research Center for Multimedia Software and \\
Hubei Key Laboratory of Multimedia and Network Communication Engineering, Wuhan University \\
$^{2}$ School of Cyber Science and Technology, Shenzhen Campus of Sun Yat-sen University \\
$^{3}$ Tencent Inc. \quad
$^{4}$ Beijing, China
}

\begin{document}

\twocolumn[{
    \renewcommand\twocolumn[1][]{#1}
    \maketitle
    \begin{center}
        \vspace{-0.5em}
        \includegraphics[width=0.95\textwidth]{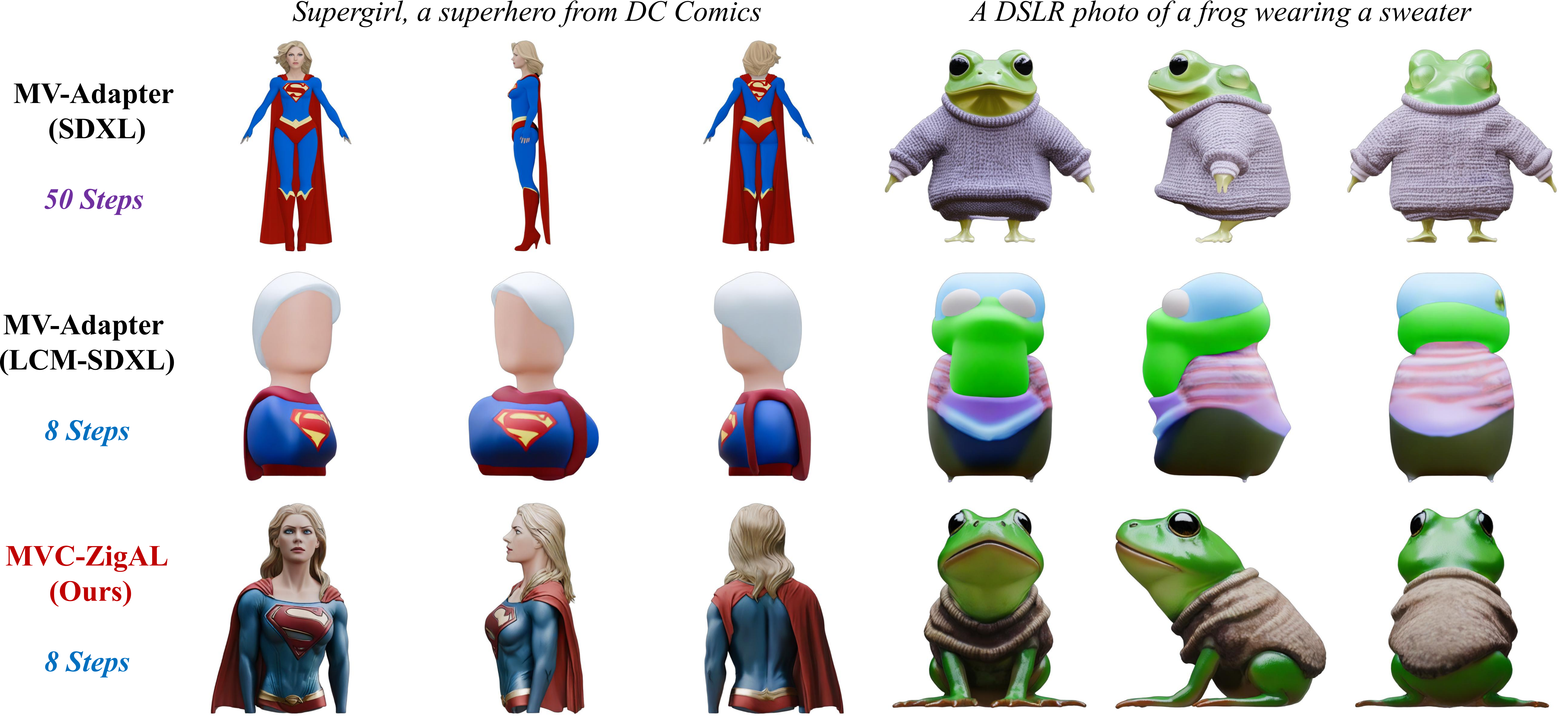}
        \captionof{figure}{\textbf{Text-to-multiview generations} from MV-Adapter (SDXL), MV-Adapter (LCM-SDXL), and our MVC-ZigAL finetuned model.}
        \label{fig:teaser}
        \vspace{1.0em}
    \end{center}
}]
\renewcommand{\thefootnote}{\fnsymbol{footnote}}
\footnotetext[1]{Corresponding authors.}
\renewcommand{\thefootnote}{\arabic{footnote}}

\begin{abstract}
Text-to-multiview (T2MV) diffusion models have shown great promise in generating multiple views of a scene from a single text prompt. While few-step backbones enable real-time T2MV generation, they often compromise key aspects of generation quality, such as per-view fidelity and cross-view consistency. Reinforcement learning (RL) finetuning offers a potential solution, yet existing approaches designed for single-image diffusion do not readily extend to the few-step T2MV setting, as they neglect cross-view coordination and suffer from weak learning signals in few-step regimes. To address this, we propose MVC-ZigAL, a tailored RL finetuning framework for few-step T2MV diffusion models. Specifically, its core insights are: (1) a new MDP formulation that jointly models all generated views and assesses their collective quality via a joint-view reward; (2) a novel advantage learning strategy that exploits the performance gains of a self-refinement sampling scheme over standard sampling, yielding stronger learning signals for effective RL finetuning; and (3) a unified RL framework that extends advantage learning with a Lagrangian dual formulation for multiview-constrained optimization, balancing single-view and joint-view objectives through adaptive primal-dual updates under a self-paced threshold curriculum that harmonizes exploration and constraint enforcement. Collectively, these designs enable robust and balanced RL finetuning for few-step T2MV diffusion models, yielding substantial gains in both per-view fidelity and cross-view consistency. Code is available at \href{https://github.com/ZiyiZhang27/MVC-ZigAL}{https://github.com/ZiyiZhang27/MVC-ZigAL}.
\end{abstract}

\section{Introduction}
\label{sec:1}


Text-to-image (T2I) diffusion models~\cite{sd,sdxl,sd3} have made rapid progress in producing high-fidelity images from text prompts. Building on these advances, \textit{text-to-multiview} (T2MV) diffusion models~\cite{mvdream,spad,viewdiff,mvadapter} jointly synthesizes multiple views of the same scene from a single text prompt. \textit{Few-step} backbones such as latent consistency models (LCMs)~\cite{lcm} unlock real-time T2MV generation by reducing model evaluations per sample to under 8. However, this direct acceleration often compromises key aspects of T2MV generation, such as image fidelity, text alignment, and cross-view consistency (see Figure~\ref{fig:teaser}, 2nd row).

Concurrently, reinforcement learning (RL) finetuning has emerged as a promising route to align T2I diffusion with human preferences or learned rewards~\cite{sftpg,dpok,ddpo,tdpo,cdpo,d3po,wallace2024diffusion,prdp,rebel,sdpo}. Yet, directly transferring these methods to the few-step T2MV setting is non-trivial, as they focus on optimizing single-image quality rewards without considering cross-view coordination, which is crucial for T2MV diffusion. Moreover, in the few-step regime, the consistently low-quality samples with rewards clustered tightly yields weak learning signals, limiting the effectiveness of standard RL finetuning. To bridge this gap, we propose \textbf{MVC-ZigAL}, a novel RL finetuning framework tailored for few-step T2MV diffusion to effectively refine both image fidelity and cross-view consistency (Figure~\ref{fig:teaser}, 3rd row).

To this end, we first reformulate T2MV diffusion as a \textit{multiview-aware} Markov decision process (MDP) that jointly models the generation of all views and employs a \textit{joint-view reward function} to evaluate their collective quality. Building on this formulation, we adapt three representative T2I RL finetuning paradigms—policy gradient~\cite{dpok,ddpo}, direct preference optimization (DPO)~\cite{d3po,wallace2024diffusion}, and reward difference learning~\cite{prdp,rebel}—to the multiview-aware MDP (Section~\ref{sec:method1}), as baselines for comparative evaluation.

Next, to address the challenges of weak learning signals in RL finetuning of few-step T2MV models, we introduce a novel \textit{advantage learning} strategy, which leverages a self-refinement sampling scheme to produce refined samples as stronger counterparts to standard sampling outputs, and then learns from their relative improvements. Thus, the key insight of this approach is to yield more informative learning signals by focusing on the structured advantages of refined samples over standard ones, rather than relying on absolute reward values, thereby facilitating more effective policy updates for few-step T2MV diffusion (Section~\ref{subsec:method2.1}).

Finally, to enable balanced and holistic optimization across both per-view and cross-view quality rewards, we design a \textit{multiview-constrained} policy optimization framework that combines the strengths of both reward types under a Lagrangian dual formulation (Section~\ref{subsec:method2.2}). The final MVC-ZigAL framework integrates advantage learning with this multiview-constrained objective and further introduces two key mechanisms: an adaptive step-size strategy that alternates between high and low values for primal-dual updates to ensure responsive yet stable constraint enforcement, and a self-paced curriculum on the constraint threshold that evolves with policy improvement, leading to more effective and stable optimization (Section~\ref{subsec:method2.3}).

Extensive experiments demonstrate that our RL finetuning framework significantly enhances few-step T2MV diffusion models across various metrics, even in out-of-distribution scenarios. Comprehensive ablation studies validate the effectiveness of each component within our framework. Our key contributions are summarized as follows:
\begin{itemize}
    \item We propose a multiview-aware MDP formulation for T2MV diffusion and adapt three representative T2I RL finetuning methods as baselines for this new setting.
    \item We introduce an advantage learning strategy that leverages the advantages of self-refined samples over standard sampling outputs to provide stronger learning signals for effective finetuning of few-step T2MV diffusion models.
    \item We develop a multiview-constrained policy optimization framework that reconciles single-view and joint-view rewards via Lagrangian duality, enhanced by adaptive primal-dual updates under a self-paced threshold curriculum balancing exploration and constraint enforcement.
\end{itemize}

\section{Preliminaries}
\label{sec:pre}

\textbf{Diffusion models}~\cite{ddpm,ddim} synthesize images through an iterative denoising process, where a neural network predicts noise residuals at each step to gradually reconstruct a clean image \(\mathbf{x}_0\) from noise \(\mathbf{x}_T\). In their practical applications to text-to-image (T2I) generation~\cite{sd,sdxl,sd3}, cross-attention layers inject text prompt embeddings \(\mathbf{c}\) into the denoising process, steering generation toward text-aligned images.

\textbf{Text-to-multiview (T2MV) diffusion models}~\cite{mvdream,spad,mvadapter} extend T2I frameworks to jointly synthesize multiple images \(\{\mathbf{x}_0^v\}_{v=1}^V\) depicting the same scene from different viewpoints under a single prompt. This is achieved by conditioning each denoising step on both the text prompt \(\mathbf{c}\) and viewpoint-specific camera embeddings \(\mathbf{e}_v\), i.e.,
\begin{equation}
    p_\theta(\mathbf{x}_{t-1}^v \mid \mathbf{x}_t^v, \mathbf{c}, \mathbf{e}_v) = \mathcal{N}\left(\mathbf{x}_{t-1} \mid \mu_\theta(\mathbf{x}_t, t, \mathbf{c}, \mathbf{e}_v), \sigma_t^2\mathbf{I}\right).
\end{equation}

Among these, MV-Adapter~\cite{mvadapter} offers a plug-and-play solution that maintains compatibility with various T2I backbones, including few-step models such as latent consistency models (LCMs)~\cite{lcm}. LCMs accelerate generation to less than 8 model evaluations (versus the usual 20–50), enabling real-time T2MV synthesis. However, this aggressive acceleration often compromises key aspects of T2MV generation quality, particularly cross-view consistency.

\textbf{RL finetuning for diffusion models}~\cite{sftpg} typically reformulates the denoising process as a Markov decision process (MDP), where each denoising step constitutes a state-action pair: \(s_t = (\mathbf{x}_t, \mathbf{c})\) and \(a_t = \mathbf{x}_{t-1}\). The RL objective of the diffusion policy is to maximize the expected reward \(\mathbb{E}[R(\mathbf{x}_0, \mathbf{c})]\) over generated images \(\mathbf{x}_0\). Various optimization paradigms have been developed within this framework, including policy gradient~\cite{dpok,ddpo}, direct preference optimization (DPO)~\cite{d3po,wallace2024diffusion}, and reward difference learning~\cite{prdp,rebel} (see Appendix~\ref{appendix:t2i_rl} for detailed formulations). However, they are inherently designed for T2I tasks, focusing on optimizing single-image rewards without considering cross-view coordination, limiting their effectiveness in T2MV.

\section{RL Finetuning for T2MV Diffusion Models}
\label{sec:method1}

\subsection{Multiview-Aware MDP Formulation}
\label{subsec:method1.1}

To address the limitations of existing T2I RL methods in T2MV, we reformulate the T2MV denoising process as a \textit{multiview-aware} MDP that jointly models the generation of all views. Unlike the original T2I MDP, which evolves a single state-action pair per step, this new formulation concurrently manages a set of \(V\) state-action pairs, each conditioned on its corresponding camera embedding \(\mathbf{e}_v\) for viewpoint \(v\). Additionally, we employ an arbitrary \textit{joint-view reward function} \(\mathcal{R}_\text{mv}\) that evaluates the collective quality of all generated views \(\{\mathbf{x}_0^v\}_{v=1}^V\). The MDP is then defined as:
\begin{align}
    & s_t \triangleq \left(\{\mathbf{x}_{t}^v, \mathbf{e}_v\}_{v=1}^V, \mathbf{c}\right), \qquad a_t \triangleq \{\mathbf{x}_{t-1}^v\}_{v=1}^V, \nonumber \\
    & \pi(a_t \mid s_t) \triangleq p_\theta\left(\{\mathbf{x}_{t-1}^v\}_{v=1}^V \mid \{\mathbf{x}_{t}^v, \mathbf{e}_v\}_{v=1}^V, \mathbf{c}\right), \nonumber \\
    & r(s_t, a_t) \triangleq
    \begin{cases}
        \mathcal{R}_\mathrm{mv}\left(\{\mathbf{x}_0^v\}_{v=1}^V, \mathbf{c}\right) & \text{if}\ t = 0 \\
        0 & \text{otherwise}
    \end{cases}.
    \label{eq:mvmdp}
\end{align}
Under this new MDP formulation, the RL objective of the T2MV diffusion policy is to maximize the expected joint-view reward over sets of multiview images, i.e.,
\begin{equation}
    \max_{\theta} \; \mathbb{E}_{\mathbf{c} \sim p(\mathbf{c})} \mathbb{E}_{\{\mathbf{x}_0^v\}_{v=1}^V \sim p_\theta\left(\cdot \mid \{\mathbf{e}_v\}_{v=1}^V, \mathbf{c}\right)} \left[ \mathcal{R}_\mathrm{mv}\left(\{\mathbf{x}_0^v\}_{v=1}^V, \mathbf{c}\right) \right].
    \label{eq:mvrl_obj}
\end{equation}

\subsection{Baselines for T2MV Policy Optimization}
\label{subsec:method1.2}

To achieve the above RL objective, we develop three policy optimization strategies within our multiview-aware MDP, as baselines for comparative evaluation with our subsequent contributions. Detailed formulations are provided below.

\textbf{(1) Multiview-aware policy gradient (MV-PG).}
This adaptation of policy gradient algorithms~\cite{dpok,ddpo} to our framework accumulates the log-likelihood gradients over all \(V\) views using the joint-view reward \(\mathcal{R}_\mathrm{mv}\). Specifically, the gradient estimator is given by:
\begin{equation}
    \mathbb{E}\left[ -\mathcal{R}_\mathrm{mv}\left(\{\mathbf{x}_0^v\}_{v=1}^V, \mathbf{c}\right) \sum_{t=1}^T \sum_{v=1}^V \nabla_\theta \log{p_\theta(\mathbf{x}_{t-1}^v \mid \mathbf{x}_{t}^v, \mathbf{e}_v, \mathbf{c})} \right].
    \label{eq:mvpg_obj}
\end{equation}

\textbf{(2) Multiview-aware direct preference optimization (MV-DPO).}
This variation of DPO~\cite{d3po,wallace2024diffusion} redefines the preference-based objective by aggregating log-likelihood ratios across all views in both preferred and non-preferred trajectories, resulting in the following MV-DPO objective:
\begin{align}
    \mathbb{E} \Bigg[ \log\sigma \Bigg( \beta \sum_{t=1}^T \sum_{v=1}^V \Bigg( \log \frac{p_{\theta} (\mathbf{x}_{t-1}^{l,v} \mid \mathbf{x}_t^{l,v}, \mathbf{e}_v, \mathbf{c})}{p_{\mathrm{ref}} (\mathbf{x}_{t-1}^{l,v} \mid \mathbf{x}_t^{l,v}, \mathbf{e}_v, \mathbf{c})} \nonumber \\
    - \log \frac{p_{\theta} (\mathbf{x}_{t-1}^{w,v} \mid \mathbf{x}_t^{w,v}, \mathbf{e}_v, \mathbf{c})}{p_{\mathrm{ref}} (\mathbf{x}_{t-1}^{w,v} \mid \mathbf{x}_t^{w,v}, \mathbf{e}_v, \mathbf{c})} \Bigg) \Bigg) \Bigg].
    \label{eq:mvdpo_obj}
\end{align}

\textbf{(3) Multiview-aware reward difference learning (MV-RDL).}
This extension of reward difference learning~\cite{prdp,rebel} minimizes the squared error between the log-likelihood ratio differences and the corresponding joint-view reward differences across all views for each pair of trajectories:
\begin{align}
    \mathbb{E} & \Bigg[ \sum_{t=1}^T \sum_{v=1}^V \Bigg( \frac{1}{\eta} \Bigg( \log \frac{p_\theta(\mathbf{x}_{t-1}^{a,v} \mid \mathbf{x}_t^{a,v}, \mathbf{e}_v, \mathbf{c})}{p_{\theta'}(\mathbf{x}_{t-1}^{a,v} \mid \mathbf{x}_t^{a,v}, \mathbf{e}_v, \mathbf{c})} \nonumber \\
    & - \log \frac{p_\theta(\mathbf{x}_{t-1}^{b,v} \mid \mathbf{x}_t^{b,v}, \mathbf{e}_v, \mathbf{c})}{p_{\theta'}(\mathbf{x}_{t-1}^{b,v} \mid \mathbf{x}_t^{b,v}, \mathbf{e}_v, \mathbf{c})} \Bigg) \label{eq:mvrdl_obj} \\
    & - \bigg( \mathcal{R}_\mathrm{mv}\left(\{\mathbf{x}_0^{a,v}\}_{v=1}^V, \mathbf{c}\right) - \mathcal{R}_\mathrm{mv}\left(\{\mathbf{x}_0^{b,v}\}_{v=1}^V, \mathbf{c}\right) \bigg) \Bigg)^2 \Bigg]. \nonumber
\end{align}

\textbf{Joint-view reward function.}
A critical component of our multiview-aware MDP is the joint-view reward function \(\mathcal{R}_\mathrm{mv}\). In this work, we use HyperScore~\cite{mate3d}, a multi-dimensional quality evaluator developed for text-to-3D generation, to compose our \(\mathcal{R}_\mathrm{mv}\). Although the complete HyperScore pipepline involves rendering multiview images from the generated 3D assets, we adapt it to our T2MV setting by directly feeding the generated multiview images into HyperScore's base scorer model, which outputs four dimensions of quality scores: \textit{alignment}, \textit{geometry}, \textit{texture}, and \textit{overall}. Among these, we use the overall quality score to construct our joint-view reward function \(\mathcal{R}_\mathrm{mv}\), while leaving the other dimensions for out-of-domain evaluation.

\section{MVC-ZigAL: A Tailored RL Finetuning Framework for Few-Step T2MV Diffusion}
\label{sec:method2}

\subsection{Learning from Self-Reflected Advantages for Few-Step T2MV Diffusion Models}
\label{subsec:method2.1}

\begin{figure*}[t]
    \centering
    \includegraphics[width=0.95\textwidth]{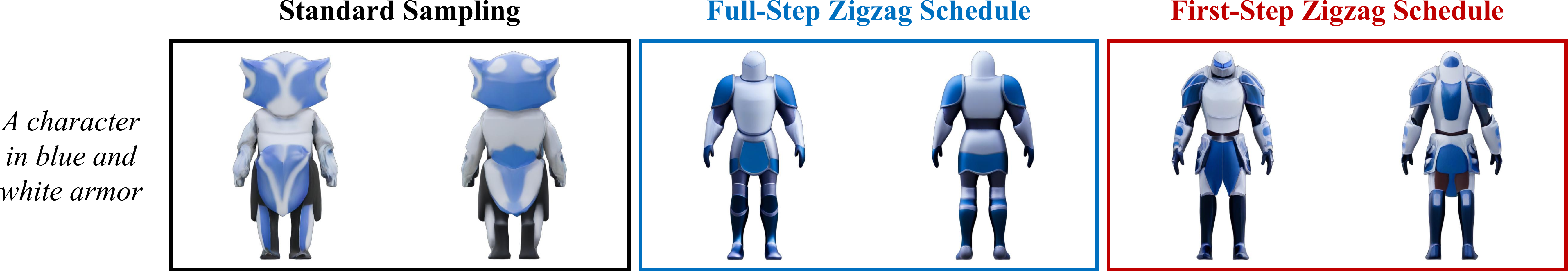}
    \caption{Comparison of \textbf{full-step and first-step zigzag schedules for ZMV-Sampling} using the non-finetuned MV-Adapter.}
    \label{fig:zigzag}
\end{figure*}

\textbf{Self-reflected refinement for T2MV diffusion.}
To address the weak learning signals that commonly arise when finetuning few-step T2MV diffusion models, we first introduce \textit{zigzag multiview sampling} (ZMV-Sampling), a self-reflection-based technique motivated by zigzag diffusion~\cite{zsampling}, to refine T2MV generation at test time and provide stronger learning references. The key insight of this technique is to reinforce viewpoint conditioning during T2MV denoising. Concretely, we perform a three-step inversion-denoising pass (termed as ``zigzag'' pass) at selected sampling steps, alternating between high-guidance denoising steps and a low-guidance inversion step, as follows:
\begin{align}
    \{\mathbf{x}_{t-1}^v\}_{v=1}^V & \sim p_\theta\left(\cdot \mid \{\mathbf{x}_{t}^v, \mathbf{e}_v\}_{v=1}^V, \mathbf{c}; \omega_{\mathrm{high}}\right), \label{eq:zigzag_a} \\
    \{\tilde{\mathbf{x}}_t^v\}_{v=1}^V & \sim q_\theta\left(\cdot \mid \{\mathbf{x}_{t-1}^v, \mathbf{e}_v\}_{v=1}^V, \mathbf{c}; \omega_{\mathrm{low}}\right), \label{eq:zigzag_b} \\
    \{\mathbf{x}_{t-1}^v\}_{v=1}^V & \sim p_\theta\left(\cdot \mid \{\tilde{\mathbf{x}}_{t}^v, \mathbf{e}_v\}_{v=1}^V, \mathbf{c}; \omega_{\mathrm{high}}\right), \label{eq:zigzag_c}
\end{align}
where \(\omega_{\mathrm{high}}\) and \(\omega_{\mathrm{low}}\) denote high and low guidance scales, respectively, applied to both text and viewpoint conditioning (while prior work only modulates text guidance), and \(q_\theta\) represents an approximate inversion operator that reintroduces noise under low guidance (see Appendix~\ref{appendix:inversion} for details). This zigzag pass with a guidance gap enables ``self-reflection'' on the initial denoising outcome, reinforcing condition-aligned features that survive the low-guidance inversion and suppressing misaligned ones during the subsequent re-denoising, thereby yielding stronger alignment.

\textbf{First-step zigzag schedule.}
While the zigzag pass can be applied at any sampling step \(t\), we find that restricting it to the very first step (\(t = T\)) yields optimal performance for T2MV generation. This design is grounded in the hierarchical nature of diffusion denoising: early steps establish global structures, while later steps refine local details and textures. Thus, applying zigzag at \(t = T\) maximizes its impact on the foundational geometric layout, whereas repeating it at later steps may interrupt the flow of detail refinement but offers diminishing returns for geometry. As shown in Figure~\ref{fig:zigzag}, the first-step zigzag schedule establishes an enhanced structural prior without restricting later refinements, yielding both enhanced geometry and finer details, whereas the full-step schedule tends to over-smooth textures.

\textbf{Multiview-aware zigzag advantage learning (MV-ZigAL).}
Leveraging self-refined trajectories from ZMV-Sampling, we then propose \textit{MV-ZigAL}, a novel advantage learning strategy tailored for few-step T2MV diffusion finetuning. Specifically, MV-ZigAL takes the reward advantages of the self-refined trajectories over standard trajectories as learning signals to guide policy updates. Given a pair of trajectories generated for the same \(\mathbf{c}\) (one via ZMV-Sampling, denoted \(\{\mathbf{x}_0^{z,v}\}_{v=1}^V\), and the other via standard sampling, denoted \(\{\mathbf{x}_0^{s,v}\}_{v=1}^V\)), we define the difference in their joint-view rewards as a \textit{zigzag advantage} function:
\begin{equation}
    \mathcal{A}_\mathrm{mv} \triangleq \mathcal{R}_\mathrm{mv}\left( \{\mathbf{x}_0^{z,v}\}_{v=1}^V, \mathbf{c} \right) - \mathcal{R}_\mathrm{mv}\left( \{\mathbf{x}_0^{s,v}\}_{v=1}^V, \mathbf{c} \right).
    \label{eq:advantage}
\end{equation}
Building on this formulation, we define our MV-ZigAL objective to minimize the squared error between the log-likelihood ratio gap of each trajectory pair and its corresponding zigzag advantage across all views, as follows:
\begin{align}
    \mathbb{E} & \Bigg[ \sum_{t=1}^T \sum_{v=1}^V \Bigg( \frac{1}{\eta} \Bigg( \log \frac{p_\theta(\mathbf{x}_{t-1}^{z,v} \mid \mathbf{x}_t^{z,v}, \mathbf{e}_v, \mathbf{c})}{p_{\theta'}(\mathbf{x}_{t-1}^{z,v} \mid \mathbf{x}_t^{z,v}, \mathbf{e}_v, \mathbf{c})} \nonumber \\
    & - \log \frac{p_\theta(\mathbf{x}_{t-1}^{s,v} \mid \mathbf{x}_t^{s,v}, \mathbf{e}_v, \mathbf{c})}{p_{\theta'}(\mathbf{x}_{t-1}^{s,v} \mid \mathbf{x}_t^{s,v}, \mathbf{e}_v, \mathbf{c})} \Bigg) - \mathcal{A}_\mathrm{mv} \Bigg)^2 \Bigg].
    \label{eq:mvzigal_obj}
\end{align}
\textbf{Discussion.}
This advantage learning strategy effectively amplifies learning signals by focusing on relative improvements between self-refined and standard trajectories, rather than absolute reward values that may be tightly clustered in few-step regimes. Compared to MV-RDL in Eq.~(\ref{eq:mvrdl_obj}), which learns from reward differences between two standard trajectories, MV-ZigAL leverages the structured advantages of self-refined trajectories from ZMV-Sampling as reference, yielding more informative gradients for policy updates.

\subsection{Multiview-Constrained Policy Optimization}
\label{subsec:method2.2}

\begin{figure*}[t]
    \centering
    \includegraphics[width=0.95\textwidth]{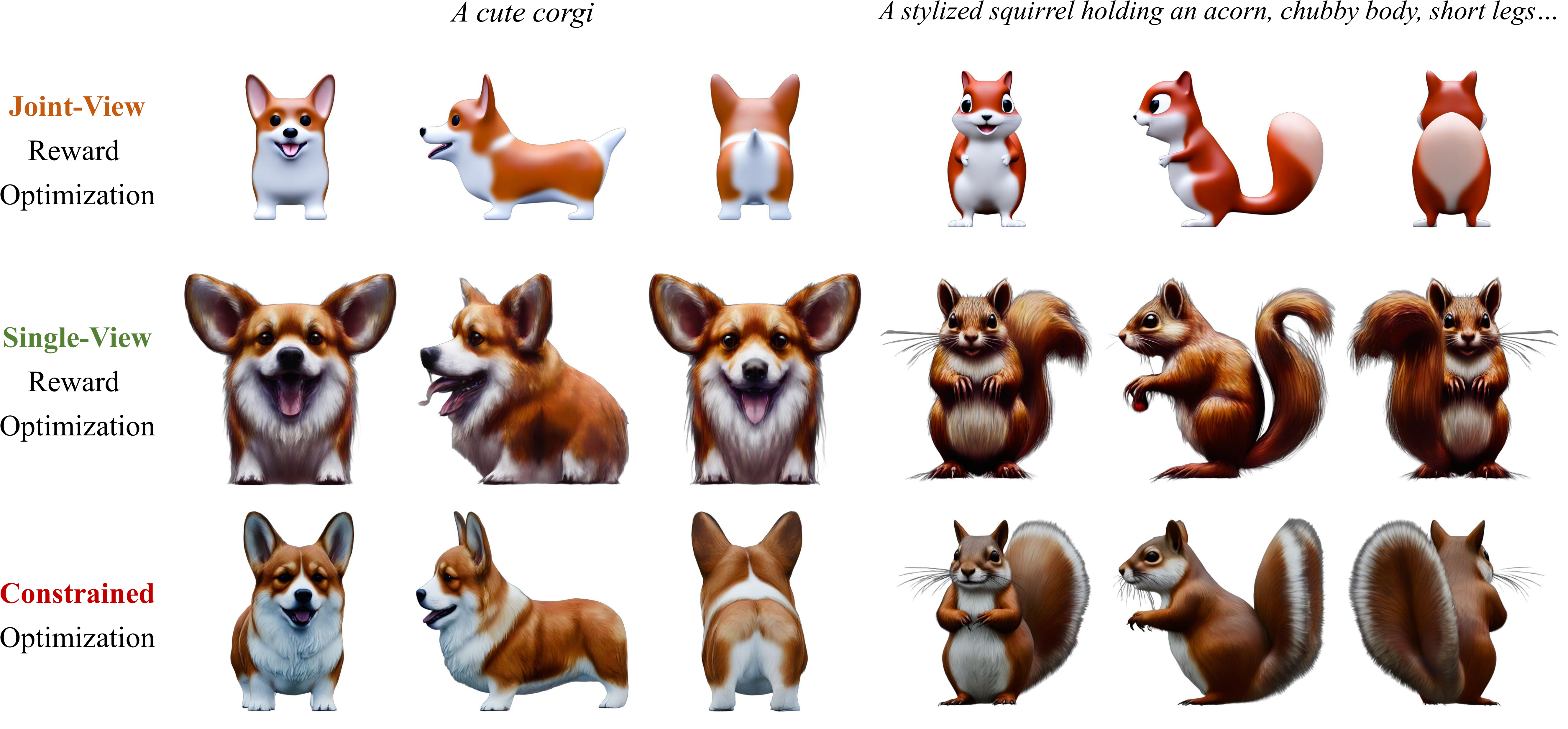}
    \caption{\textbf{Qualitative comparison of reward optimization} with joint-view or single-view rewards only versus our multiview-constrained approach. \textit{Joint-view optimization} emphasizes view consistency but under-optimizes image fidelity; \textit{single-view optimization} targets image fidelity but compromises view consistency, causing the ``multi-face'' problem. In contrast, our approach balances both aspects effectively.}
    \label{fig:constrained}
\end{figure*}

\textbf{Joint-view rewards vs. single-view rewards.}
While joint-view rewards effectively evaluate the collective quality of all generated views, they may lack the granularity needed to provide detailed feedback for each individual view, potentially leading to suboptimal image fidelity. Conversely, single-view rewards for T2I tasks, such as PickScore~\cite{pick} and HPSv2~\cite{hpsv2}, offer fine-grained feedback for each view, but may neglect cross-view coordination, resulting in inconsistent multiview outputs (e.g., the ``multi-face'' problem illustrated in Figure~\ref{fig:constrained}). To address this, we propose a multiview-constrained policy optimization framework that synergistically combines the strengths of both reward types.

\textbf{Multiview-constrained MDP formulation.}
We first redefine our multiview-aware MDP from Eq.~(\ref{eq:mvmdp}) to incorporate single-view rewards \(R(\mathbf{x}_0^v, \mathbf{c})\) for each view \(v\) as primary rewards, while enforcing a lower bound constraint on the expected joint-view reward \(\mathcal{R}_\mathrm{mv}\). This reformulation yields a \textit{multiview-constrained policy optimization} problem, transforming the RL objective in Eq.~(\ref{eq:mvrl_obj}) as follows:
\begin{align}
    \max_{\theta} \;\;\; & \mathbb{E}_{\mathbf{c} \sim p(\mathbf{c})} \mathbb{E}_{\{\mathbf{x}_0^v\}_{v=1}^V \sim p_\theta\left(\cdot \mid \{\mathbf{e}_v\}_{v=1}^V, \mathbf{c}\right)} \left[ \sum_{v=1}^V R(\mathbf{x}_0^v, \mathbf{c}) \right], \nonumber \\
    \mathrm{s.t.} \;\;\; & \mathbb{E} \left[ \mathcal{R}_\mathrm{mv}\left(\{\mathbf{x}_0^v\}_{v=1}^V, \mathbf{c}\right) \right] \geq \tau,
    \label{eq:constrained_obj}
\end{align}
where \(\tau\) is a predefined threshold that sets the minimum acceptable level of the expected joint-view reward. This multiview-constrained formulation explicitly balances the optimization of individual view fidelity with the enforcement of cross-view consistency, as illustrated in Figure~\ref{fig:constrained}.

\textbf{Lagrangian dual optimization.}
We solve this constrained optimization problem via the Lagrangian dual approach~\cite{cpo}, introducing a Lagrange multiplier \(\lambda \geq 0\) to incorporate the joint-view constraint into the objective, i.e.,
\begin{equation}
    \min_{\lambda \geq 0} \max_\theta \; \mathbb{E} \left[ \sum_{v=1}^V R(\mathbf{x}_0^v, \mathbf{c}) + \lambda \left( \mathcal{R}_\mathrm{mv}\left(\{\mathbf{x}_0^v\}_{v=1}^V, \mathbf{c}\right) - \tau \right) \right].
    \label{eq:lagrangian}
\end{equation}

\textbf{Multiview-constrained reward function.}
To streamline the dual optimization process, we define a \textit{multiview-constrained reward function} that amalgamates single-view and joint-view rewards into a unified signal for each view:
\begin{equation}
    \mathcal{R}_\mathrm{mvc}(\mathbf{x}_0^v, \mathbf{c}) = \frac{R(\mathbf{x}_0^v, \mathbf{c}) + \lambda \mathcal{R}_\mathrm{mv}\left(\{\mathbf{x}_0^v\}_{v=1}^V, \mathbf{c}\right)}{1 + \lambda},
    \label{eq:c_reward}
\end{equation}
which balances the contributions of both reward types based on the current value of \(\lambda\). This formulation allows us to treat multiview-constrained optimization as a standard RL problem with modified rewards, facilitating the application of our previously developed advantage learning strategy.

\subsection{Adaptive Primal-Dual Updates under a Self-Paced Constraint Curriculum}
\label{subsec:method2.3}

\textbf{Multiview-constrained advantage learning.}
To implement our advantage learning strategy within the multiview-constrained framework, we reformulate the zigzag advantage function (Eq.~\eqref{eq:advantage}) by replacing the joint-view rewards with the multiview-constrained rewards from Eq.~(\ref{eq:c_reward}), yielding this \textit{multiview-constrained advantage} function:
    \begin{equation}
    \mathcal{A}_\mathrm{mvc}(\mathbf{x}_0^{z,v}, \mathbf{x}_0^{s,v},\mathbf{c}) = \mathcal{R}_\mathrm{mvc}(\mathbf{x}_0^{z,v}, \mathbf{c}) - \mathcal{R}_\mathrm{mvc}(\mathbf{x}_0^{s,v}, \mathbf{c}),
    \label{eq:c_advantage}
\end{equation}
where \(\mathcal{R}_\mathrm{mvc}(\mathbf{x}_0^{z,v}, \mathbf{c})\) and \(\mathcal{R}_\mathrm{mvc}(\mathbf{x}_0^{s,v}, \mathbf{c})\) are the multiview-constrained rewards for the ZMV-Sampling and standard sampling trajectories, respectively. This leads to our final \textit{MVC-ZigAL} objective, which minimizes the squared error between the log-likelihood ratio gap of each trajectory pair and its multiview-constrained advantage across all views:
\begin{align}
    \mathbb{E} & \Bigg[ \sum_{t=1}^T \sum_{v=1}^V \Bigg( \frac{1}{\eta} \Bigg( \log \frac{p_\theta(\mathbf{x}_{t-1}^{z,v} \mid \mathbf{x}_t^{z,v}, \mathbf{e}_v, \mathbf{c})}{p_{\theta'}(\mathbf{x}_{t-1}^{z,v} \mid \mathbf{x}_t^{z,v}, \mathbf{e}_v, \mathbf{c})} \nonumber \\
    & - \log \frac{p_\theta(\mathbf{x}_{t-1}^{s,v} \mid \mathbf{x}_t^{s,v}, \mathbf{e}_v, \mathbf{c})}{p_{\theta'}(\mathbf{x}_{t-1}^{s,v} \mid \mathbf{x}_t^{s,v}, \mathbf{e}_v, \mathbf{c})} \Bigg) - \mathcal{A}_\mathrm{mvc}(\mathbf{x}_0^{z,v}, \mathbf{x}_0^{s,v},\mathbf{c}) \Bigg)^2 \Bigg].
    \label{eq:mvczigal_obj}
\end{align}

\textbf{Primal-dual updates with adaptive step size.}
To effectively enforce the joint-view constraint within MVC-ZigAL, we employ a primal-dual update scheme that concurrently optimizes the policy parameters \(\theta\) and the Lagrange multiplier \(\lambda\) during training. At each training iteration \(k\), before updating the policy, we first compute the average joint-view reward across a batch of \(B\) trajectory pairs:
\begin{equation}
    \bar{\mathcal{R}}_\mathrm{mv} = \frac{1}{2B} \sum_{i=1}^B \left( \mathcal{R}_\mathrm{mv}^{i,s} + \mathcal{R}_\mathrm{mv}^{i,z} \right),
    \label{eq:avg_reward}
\end{equation}
where \(\mathcal{R}_\mathrm{mv}^{i,s}\) and \(\mathcal{R}_\mathrm{mv}^{i,z}\) denote the joint-view rewards for the \(i\)-th pairs of standard and ZMV-Sampling trajectories, respectively. We then update \(\lambda\) in proportion to the deviation of \(\bar{\mathcal{R}}_\mathrm{mv}\) from the threshold \(\tau\), using an adaptive step size:
\begin{equation}
    \lambda_{k+1} \leftarrow \max\left( \lambda_k + \alpha^{\pm} \left( \tau - \bar{\mathcal{R}}_\mathrm{mv} \right), 0 \right),
    \label{eq:lambda_update}
\end{equation}
where \(\alpha^{\pm}\) is the adaptive step size depending on whether the constraint is satisfied or violated. Specifically, we apply a larger step size \(\alpha^{+}\) upon violation (\(\bar{\mathcal{R}}_\mathrm{mv} < \tau\)) to promptly reinforce the constraint, and a smaller step size \(\alpha^{-}\) upon satisfaction (\(\bar{\mathcal{R}}_\mathrm{mv} \geq \tau\)) to allow smoother, more stable adjustments that reduce oscillations around the threshold. This adaptive strategy ensures responsive yet stable enforcement of the joint-view constraint throughout training.

\textbf{Self-paced constraint thresholding.}
Since a fixed threshold \(\tau\) may be too stringent or lenient at different training stages, either hindering exploration or weakening constraint enforcement, we further propose a \textit{self-paced curriculum} for the threshold \(\tau\) to dynamically adapt it based on the policy's evolving performance. Specifically, we update \(\tau\) using an exponential moving average (EMA) of \(\bar{\mathcal{R}}_\mathrm{mv}\):
\begin{equation}
    \tau_{k+1} \leftarrow \beta_\tau \tau_{k} + (1 - \beta_\tau) \bar{\mathcal{R}}_\mathrm{mv},
\end{equation}
where \(\beta_\tau \in [0, 1)\) is the EMA smoothing factor. This self-paced curriculum allows the constraint to evolve in tandem with the policy's learning dynamic, encouraging broader exploration in early stages and progressively tightening the constraint as the policy improves. Accordingly, it can preemptively react to emerging regressions in joint-view rewards, preventing prolonged constraint breaches rather than merely responding after violations occur.

\section{Experiments}
\label{sec:exp}

\subsection{Experimental Setups}
\label{subsec:setup}

In our main experiments, we use MV-Adapter~\cite{mvadapter} with an LCM-SDXL backbone as the few-step T2MV baseline, and apply LoRA~\cite{lora} to both the MV-Adapter and LCM-SDXL layers for efficient finetuning. We conduct RL finetuning using 8 sampling steps and 6 views, along with a common prompt set of 45 animal names used in prior works~\cite{ddpo,d3po,rebel} to ensure consistency with their established studies. Additionally, we adopt the MATE-3D~\cite{mate3d} benchmark, which contains 160 text prompts covering a wide range of object categories and compositions, to assess the generalization of finetuned models to unseen, complex prompts.

\textbf{Reward functions.}
We employ PickScore~\cite{pick} and HPSv2~\cite{hpsv2} as single-view reward functions to drive policy improvements in per-view image fidelity, while HyperScore \textit{overall quality}~\cite{mate3d} as the joint-view reward function focusing on cross-view consistency, as detailed in Section~\ref{subsec:method1.2}.

\textbf{Evaluation metrics.}
We use the above reward functions as primary metrics to evaluate reward optimization performance. Additionally, we report HyperScore's subcategories (\textit{alignment}, \textit{geometry}, and \textit{texture}) and ImageReward~\cite{imagereward}, to provide out-of-domain assessments of finetuned models. Note that we do not explicitly compare inference speed since RL finetuning merely modifies model weights without changing the sampling procedure, thus maintaining the same inference speed as the base model across all methods.

\begin{table}[t]
    \caption{\textbf{Quantitative comparison of T2MV policy optimization methods} using HyperScore and PickScore. All reward scores are evaluated with 8 sampling steps and 6 views per trajectory.}
    \label{tab:mvrl}
    \centering
    \resizebox{\linewidth}{!}{
    \begin{tabular}{lccccc}
    \toprule
    \textbf{Method} & \multicolumn{4}{c}{\textbf{HyperScore}} & \textbf{PickScore} \\
    & Alignment & Geometry & Texture & Overall & \\
    \midrule
    Baseline & 7.60 & 7.55 & 6.46 & 7.23 & 0.196 \\
    MV-PG & 8.49 & 9.11 & 7.21 & 8.39 & 0.203 \\
    MV-DPO & 8.07 & 8.75 & 7.02 & 8.00 & 0.200 \\
    MV-RDL & 8.90 & 9.87 & 7.80 & 9.03 & 0.203 \\
    \textbf{MV-ZigAL} & \textbf{9.20} & \textbf{9.92} & \textbf{7.84} & \textbf{9.17} & \textbf{0.205} \\
    \bottomrule
    \end{tabular}
    }
\end{table}

\begin{figure*}[t]
    \centering
    \hfill
    \includegraphics[width=0.29\textwidth]{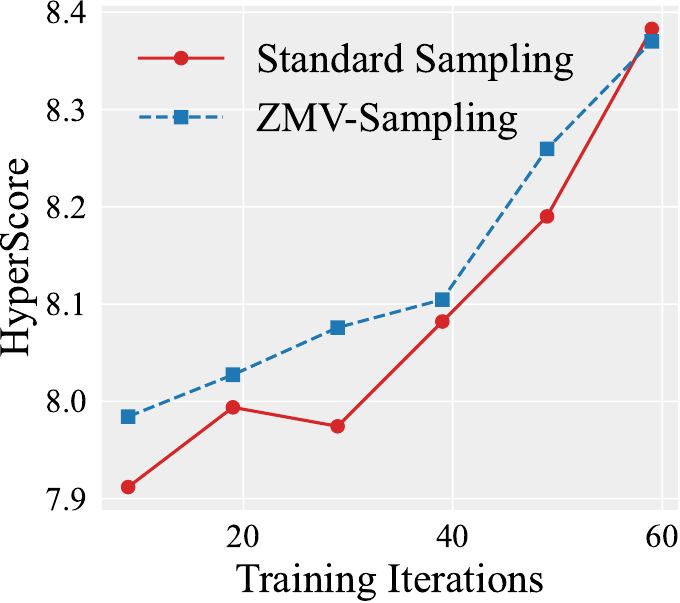}
    \hfill
    \includegraphics[width=0.3\textwidth]{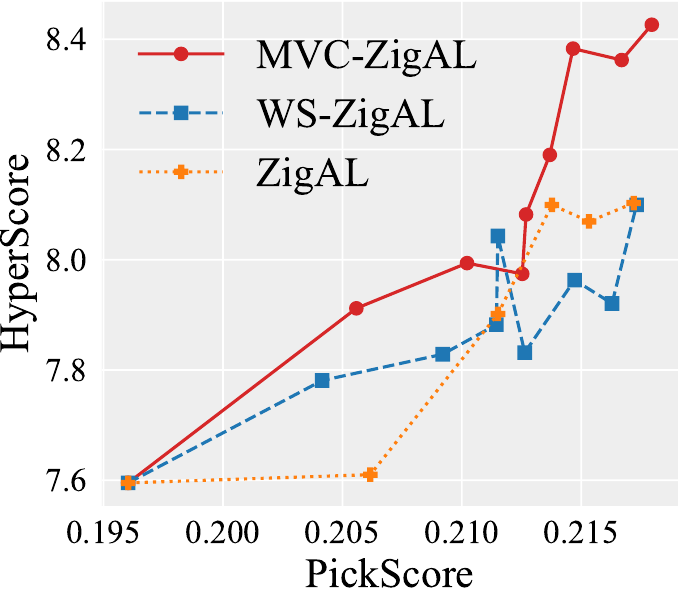}
    \hfill
    \includegraphics[width=0.3\textwidth]{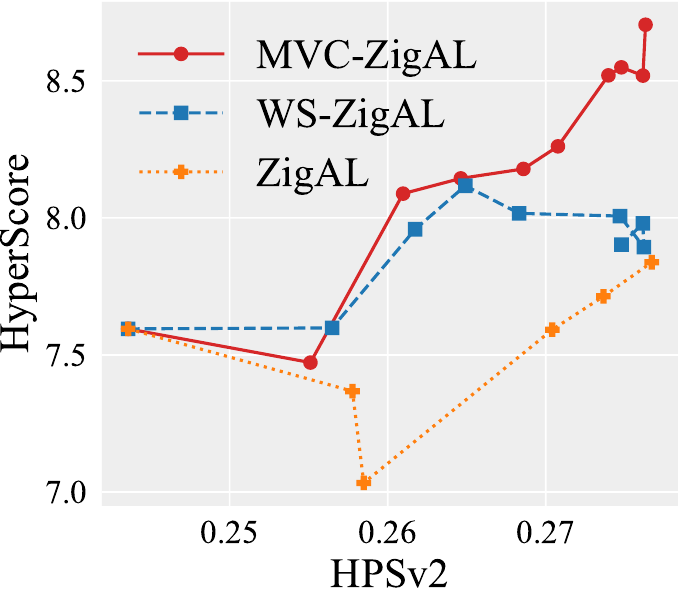}
    \hfill
    \caption{\textbf{(Left) Reward gap} in HyperScore between standard sampling and ZMV-Sampling over MVC-ZigAL finetuning. \textbf{(Middle \& Right) Trade-off} between per-view rewards (PickScore/HPSv2) and joint-view rewards (HyperScore) under different RL paradigms.}
    \label{fig:plot}
\end{figure*}

\begin{table*}[t]
    \caption{\textbf{Quantitative comparison of MVC-ZigAL variants and ablations.} All reward scores are evaluated on unseen prompts from the MATE-3D benchmark using 8 sampling steps and 6 views per trajectory, with models checkpointed at the 70th training epoch.}
    \label{tab:ablation}
    \centering
    \resizebox{0.95\linewidth}{!}{
    \begin{tabular}{lccccccc}
    \toprule
    \textbf{Method} & \multicolumn{4}{c}{\textbf{HyperScore}} & \textbf{PickScore} & \textbf{HPSv2} & \textbf{ImageReward} \\
    & Alignment & Geometry & Texture & Overall \\
    \midrule
    Baseline & 6.69 & 6.97 & 6.54 & 6.67 & 0.204 & 0.252 & -0.846 \\
    MV-ZigAL & 6.88 & 7.41 & 6.69 & 6.95 & 0.205 & 0.254 & -0.770 \\
    WS-ZigAL (\(w_\mathrm{mv} = 0.1\)) & 6.22 & 6.81 & 5.98 & 6.25 & 0.220 & 0.265 & 0.020 \\
    WS-ZigAL (\(w_\mathrm{mv} = 0.5\)) & 6.78 & 7.23 & 6.69 & 6.83 & 0.217 & 0.270 & 0.183 \\
    WS-ZigAL (\(w_\mathrm{mv} = 1.0\)) & 6.87 & 7.19 & 6.47 & 6.68 & 0.215 & 0.262 & 0.006 \\
    MVC-ZigPG (First-Step Zigzag) & 6.79 & 6.25 & 6.74 & 6.86 & 0.216 & 0.263 & 0.083 \\
    MVC-ZigAL (Full-Step Zigzag) & 6.81 & 7.33 & 6.77 & 6.91 & 0.217 & 0.270 & 0.211 \\
    MVC-ZigAL (First-Step Zigzag) & 7.01 & 7.39 & 6.89 & 7.04 & 0.217 & 0.268 & 0.180 \\
    \bottomrule
    \end{tabular}
    }
\end{table*}

\subsection{Quantitative Analysis}
\label{subsec:results}

\textbf{Comparison with T2MV policy optimization baselines.}
We first compare our MV-ZigAL against three T2MV policy optimization baselines introduced in Section~\ref{subsec:method1.2}: MV-PG, MV-DPO, and MV-RDL. As shown in Table~\ref{tab:mvrl}, our MV-ZigAL outperforms all baselines across all HyperScore metrics and achieves a slight advantage in PickScore, demonstrating its superior capability in enhancing T2MV generation quality through effective advantage learning.

\textbf{Effectiveness of advantage learning.}
To validate the effectiveness of our advantage learning strategy in internalizing ZMV-Sampling's advantages into the base model, we track the reward gap in HyperScore between standard sampling and ZMV-Sampling over MVC-ZigAL finetuning, as illustrated in the left plot of Figure~\ref{fig:plot}. Initially, ZMV-Sampling yields a clear advantage over standard sampling, but this gap narrows progressively and eventually converges, indicating a successful advantage transfer from ZMV-Sampling to the base model through MVC-ZigAL.

\textbf{Multiview-constrained policy optimization.}
To assess the efficacy of our multiview-constrained optimization framework in balancing per-view fidelity and cross-view consistency, we compare MVC-ZigAL against two alternative RL paradigms: \textbf{(1)} ZigAL, which relies solely on single-view rewards (PickScore or HPSv2) for optimization, and \textbf{(2)} WS-ZigAL, which employs a weighted sum of single-view and joint-view rewards. As shown in the middle and right plots of Figure~\ref{fig:plot}, both alternatives exhibit a trade-off between per-view and joint-view rewards, lagging in HyperScore when excelling in PickScore or HPSv2. In contrast, our MVC-ZigAL consistently achieves higher joint-view rewards as per-view rewards improve, demonstrating its effectiveness in harmonizing both objectives.

\subsection{Qualitative Analysis}
\label{subsec:vis}

Figure~\ref{fig:teaser} compares T2MV generations from the baseline MV-Adapter (with either SDXL or LCM-SDXL) and our MVC-ZigAL finetuned model with 6 views per prompt (yet we only visualize 3 views here for brevity). For a fair comparison, all visual results presented in this paper are generated using the same random seed (42). As shown, our MVC-ZigAL finetuned model produces multiview images with significantly enhanced geometric consistency and finer details compared to the few-step baseline, closely matching the quality of the computationally intensive 50-step baseline. Additional qualitative comparisons, including generations with fewer sampling steps or from other T2MV baselines, such as SPAD~\cite{spad}, are provided in Appendix~\ref{appendix:vis}.

\subsection{Ablation Study}
\label{subsec:ablation}

\begin{figure*}[ht]
    \centering
    \includegraphics[width=0.33\textwidth]{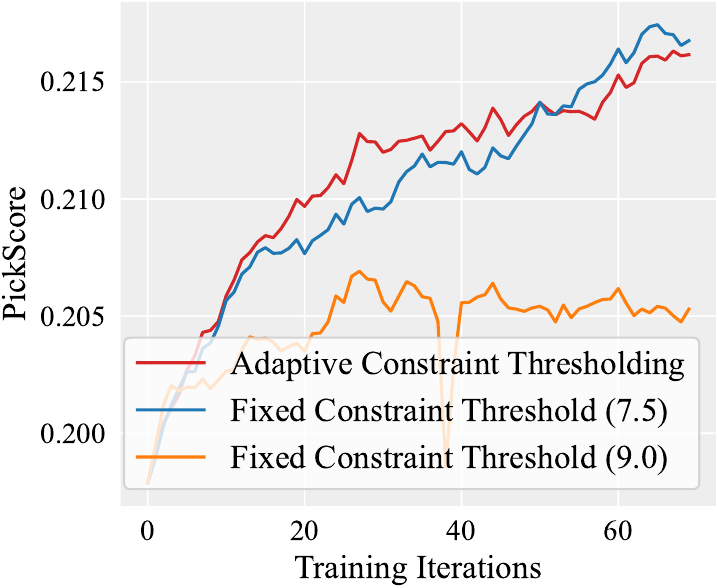}
    \hfill
    \includegraphics[width=0.32\textwidth]{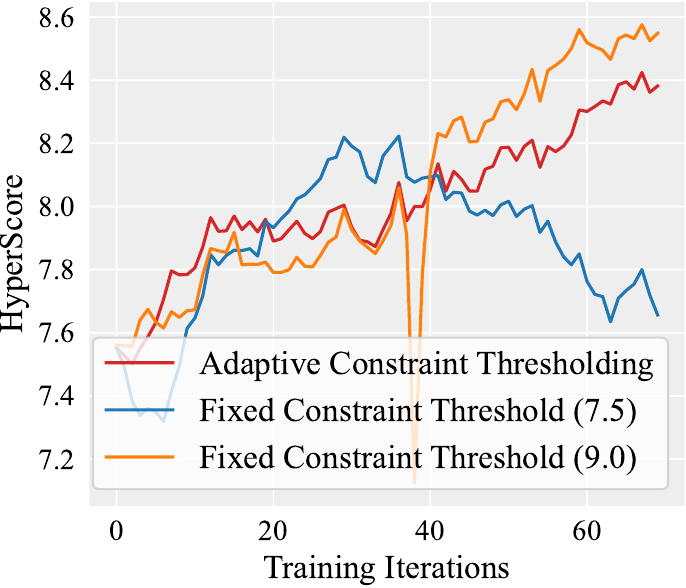}
    \hfill
    \includegraphics[width=0.31\textwidth]{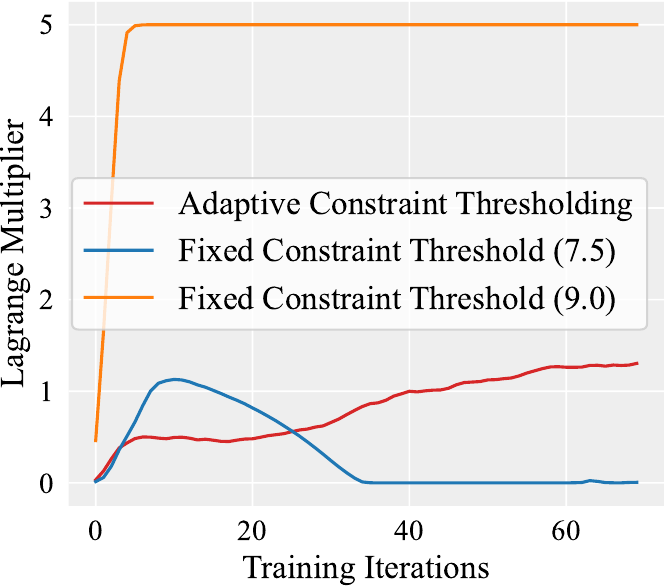}
    \caption{\textbf{Comparison of reward curves (left \& middle) and Lagrange multiplier dynamics (right)} during training of MVC-ZigAL variants with either adaptive or fixed constraint thresholds.}
    \label{fig:thr}
\end{figure*}

\begin{figure*}[ht]
    \centering
    \includegraphics[width=0.32\textwidth]{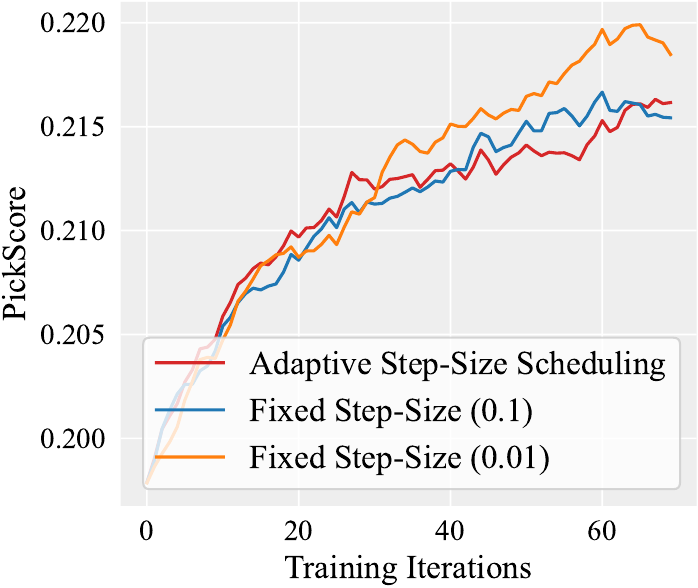}
    \hfill
    \includegraphics[width=0.31\textwidth]{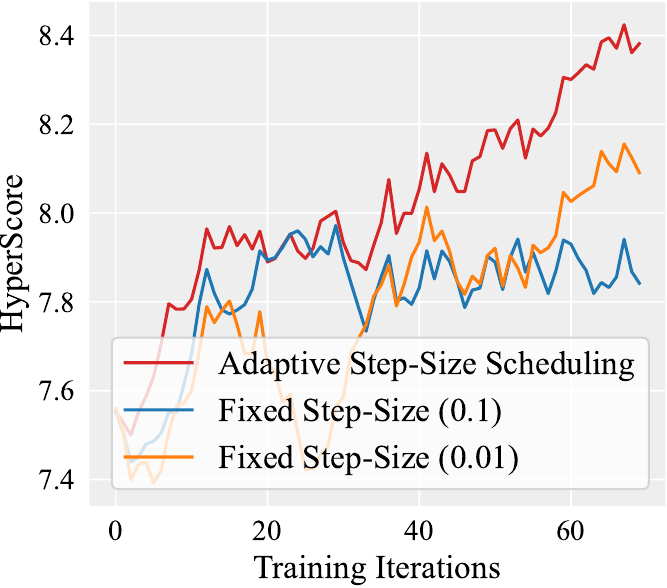}
    \hfill
    \includegraphics[width=0.32\textwidth]{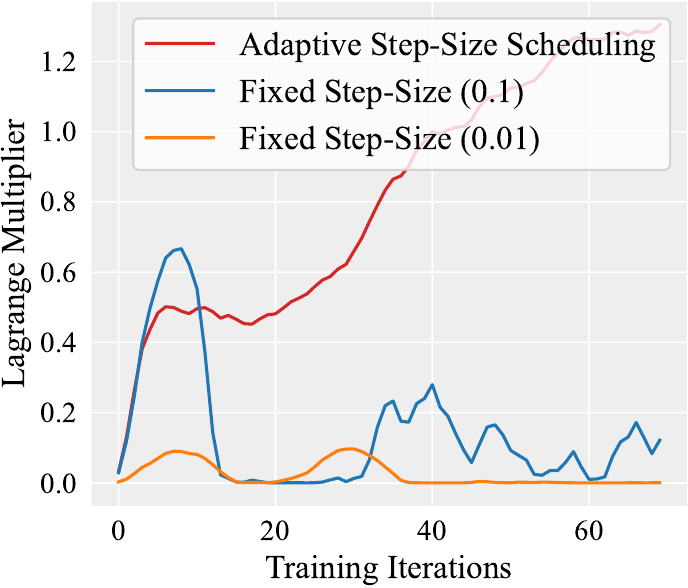}
    \caption{\textbf{Comparison of reward curves (left \& middle) and Lagrange multiplier dynamics (right)} during training of MVC-ZigAL variants using either adaptive or fixed step sizes for Lagrange multiplier updates.}
    \label{fig:stepsize}
\end{figure*}

\textbf{Effect of zigzag pass in advantage learning.}
To validate the specific contribution of the zigzag pass in our advantage learning framework, we conduct an ablation study comparing MVC-ZigAL against MVC-ZigPG, which employs standard policy gradients instead of advantage learning while retaining the zigzag pass for trajectory sampling. As shown in Table~\ref{tab:ablation}, MVC-ZigAL outperforms MVC-ZigPG across all reward metrics, demonstrating the effectiveness of our advantage learning strategy in better leveraging the self-refined trajectories from ZMV-Sampling.

\textbf{First-step zigzag vs. full-step zigzag.}
To validate the design choice of first-step zigzag scheduling in ZMV-Sampling, we conduct an ablation study comparing MVC-ZigAL with a variant that applies the zigzag pass at every sampling step (full-step zigzag). As shown in Table~\ref{tab:ablation}, the first-step zigzag schedule outperforms the full-step variant in HyperScore while achieving comparable PickScore and HPSv2, eliminating the need for costly full-step zigzag.

\textbf{Weighted sum vs. constrained optimization.}
As mentioned in Section~\ref{subsec:results}, we compare our MVC-ZigAL against WS-ZigAL, which optimizes a weighted sum of single-view and joint-view rewards. As shown in Table~\ref{tab:ablation}, across different weight settings (\(w_\mathrm{mv} \in \{0.1, 0.5, 1.0\}\)), WS-ZigAL consistently underperforms MVC-ZigAL in HyperScore, despite achieving competitive single-view metrics only when \(w_\mathrm{mv} = 0.5\). This confirms that the efficacy of the weighted sum relies heavily on careful weight tuning, whereas our constrained framework effectively balances both objectives without manual weight selection.

\textbf{Fixed vs. adaptive thresholds.}
To validate the effectiveness of our self-paced curriculum for adaptive constraint thresholding, we conduct an ablation study comparing our MVC-ZigAL with two variants using fixed thresholds, either set to the initial joint-view reward of the baseline model (7.5) or to the maximum joint-view reward achieved during training (9.0). As shown in Figure~\ref{fig:thr}, the fixed thresholds either fails to enforce the joint-view constraint consistently due to being too lenient (7.5), or hinders effective improvements in single-view rewards from being too stringent (9.0). In contrast, our adaptive thresholding enables steady improvements in both single-view and joint-view rewards, highlighting its benefits in balancing exploration and constraint enforcement without manual threshold tuning.

\textbf{Fixed vs. adaptive step sizes.}
To validate the effectiveness of our adaptive step size strategy for Lagrange multiplier updates, we conduct an ablation study comparing our MVC-ZigAL with two variants using fixed step sizes, either small (0.01) or large (0.1). As shown in Figure~\ref{fig:stepsize}, the small fixed step size reacts too slowly to constraint violations, leading to prolonged breaches and suboptimal joint-view rewards. Conversely, the large fixed step size induces oscillations in the Lagrange multiplier, destabilizing training and hindering consistent constraint satisfaction. In contrast, our adaptive step size strategy facilitates responsive yet stable constraint enforcement, yielding steady progress in both single-view and joint-view rewards.

\section{Conclusion}
\label{sec:conclusion}

In this work, we presented MVC-ZigAL, a novel RL finetuning framework for few-step T2MV diffusion models. By learning self-reflected refinements via advantage learning and enforcing joint-view constraints via Lagrangian dual optimization, MVC-ZigAL effectively enhances both per-view fidelity and cross-view consistency in a balanced manner. Future work may explore extending our framework or insights to other generative tasks requiring multiview coherence, such as video generation or 3D reconstruction.

\section*{Acknowledgments}
This work is supported in part by the National Natural Science Foundation of China (Grant No.~U23A20318, 62276195 and~62376200), the Science and Technology Major Project of Hubei Province (Grant No.~2024BAB046 and~2025BCB026) and the Foundation for Innovative Research Groups of Hubei Province (Grant No.~2024AFA017). This work was also supported by WHU-Kingsoft Joint Lab. Li Shen is supported by NSFC Grant (No.~62576364), Shenzhen Basic Research Project (Natural Science Foundation) Basic Research Key Project (NO.~JCYJ20241202124430041), CCF-Tencent Rhino-Bird Open Research Fund (NO. CCF-Tencent~RAGR20250114) and Tencent~JR2025TEG002. The numerical calculations in this paper have been done on the supercomputing system in the Supercomputing Center of Wuhan University.

{
    \small
    \bibliographystyle{ieeenat_fullname}
    \bibliography{references}

@String(AAAI = {AAAI})

@article{dpok,
 author    = {Ying Fan and Olivia Watkins and Yuqing Du and Hao Liu and Moonkyung Ryu and Craig Boutilier and Pieter Abbeel and Mohammad Ghavamzadeh and Kangwook Lee and Kimin Lee},
 title     = {{DPOK}: {R}einforcement learning for fine-tuning text-to-image diffusion models},
 year      = {2023},
 journal = {Advances in Neural Information Processing Systems},
 volume    = {36},
 pages     = {79858--79885}
}

@inproceedings{ddpo,
  author    = {Kevin Black and Michael Janner and Yilun Du and Ilya Kostrikov and Sergey Levine},
  title     = {Training diffusion models with reinforcement learning},
  booktitle = {International Conference on Learning Representations},
  year      = {2024}
}

@inproceedings{d3po,
  title={Using human feedback to fine-tune diffusion models without any reward model},
  author={Yang, Kai and Tao, Jian and Lyu, Jiafei and Ge, Chunjiang and Chen, Jiaxin and Shen, Weihan and Zhu, Xiaolong and Li, Xiu},
  booktitle={Proceedings of the IEEE/CVF Conference on Computer Vision and Pattern Recognition},
  pages={8941--8951},
  year={2024}
}

@inproceedings{wallace2024diffusion,
  title={Diffusion model alignment using direct preference optimization},
  author={Wallace, Bram and Dang, Meihua and Rafailov, Rafael and Zhou, Linqi and Lou, Aaron and Purushwalkam, Senthil and Ermon, Stefano and Xiong, Caiming and Joty, Shafiq and Naik, Nikhil},
  booktitle={Proceedings of the IEEE/CVF Conference on Computer Vision and Pattern Recognition},
  pages={8228--8238},
  year={2024}
}

@inproceedings{sftpg,
 author    = {Ying Fan and Kangwook Lee},
 title     = {Optimizing {DDPM} Sampling with Shortcut Fine-Tuning},
 booktitle = {International Conference on Machine Learning},
 pages     = {9623--9639},
 year      = {2023}
}

@article{ddpm,
 author    = {Jonathan Ho and Ajay Jain and Pieter Abbeel},
 title     = {Denoising diffusion probabilistic models},
 year      = {2020},
 journal = {Advances in Neural Information Processing Systems},
 volume    = {33},
 pages     = {6840--6851}
}

@inproceedings{ddim,
 author    = {Jiaming Song and Chenlin Meng and Stefano Ermon},
 title     = {Denoising diffusion implicit models},
 year      = {2021},
 booktitle = {International Conference on Learning Representations}
}

@inproceedings{sd,
 author    = {Robin Rombach and Andreas Blattmann and Dominik Lorenz and Patrick Esser and Björn Ommer},
 title     = {High-resolution image synthesis with latent diffusion models},
 year      = {2022},
 booktitle = {Proceedings of the IEEE/CVF Conference on Computer Vision and Pattern Recognition},
 pages     = {10684--10695}
}

@inproceedings{lora,
 author    = {Edward J Hu and Yelong Shen and Phillip Wallis and Zeyuan Allen-Zhu and Yuanzhi Li and Shean Wang and Lu Wang and Weizhu Chen},
 title     = {{LoRA}: {L}ow-rank adaptation of large language models},
 year      = {2022},
 booktitle = {International Conference on Learning Representations}
}

@article{reinforce,
  title={Simple statistical gradient-following algorithms for connectionist reinforcement learning},
  author={Williams, Ronald J},
  journal={Machine learning},
  volume={8},
  pages={229--256},
  year={1992}
}

@article{ppo,
 author    = {John Schulman and Filip Wolski and Prafulla Dhariwal and Alec Radford and Oleg Klimov},
 title     = {Proximal policy optimization algorithms},
 year      = {2017},
 journal   = {arXiv preprint arXiv:1707.06347}
}

@article{aes,
 author    = {Christoph Schuhmann and Romain Beaumont and Richard Vencu and Cade W Gordon and Ross Wightman and Mehdi Cherti and Theo Coombes and Aarush Katta and Clayton Mullis and Mitchell Wortsman and Patrick Schramowski and Srivatsa R Kundurthy and Katherine Crowson and Ludwig Schmidt and Robert Kaczmarczyk and Jenia Jitsev},
 title     = {{LAION-5B}: {A}n open large-scale dataset for training next generation image-text models},
 year      = {2022},
 journal = {Advances in Neural Information Processing Systems},
 volume    = {35},
 pages     = {25278--25294}
}

@article{hpsv2,
 author    = {Xiaoshi Wu and Yiming Hao and Keqiang Sun and Yixiong Chen and Feng Zhu and Rui Zhao and Hongsheng Li},
 title     = {{Human Preference Score v2}: {A} solid benchmark for evaluating human preferences of text-to-image synthesis},
 year      = {2023},
 journal   = {arXiv preprint arXiv:2306.09341}
}

@article{pick,
 author    = {Yuval Kirstain and Adam Polyak and Uriel Singer and Shahbuland Matiana and Joe Penna and Omer Levy},
 title     = {Pick-a-Pic: {A}n Open Dataset of User Preferences for Text-to-Image Generation},
 year      = {2023},
 journal = {Advances in Neural Information Processing Systems},
 volume    = {36},
 pages     = {36652--36663}
}

@article{imagereward,
 author    = {Jiazheng Xu and
              Xiao Liu and
              Yuchen Wu and
              Yuxuan Tong and
              Qinkai Li and
              Ming Ding and
              Jie Tang and
              Yuxiao Dong},
 title     = {{ImageReward}: {L}earning and Evaluating Human Preferences for Text-to-Image Generation},
 journal = {Advances in Neural Information Processing Systems},
 volume    = {36},
 pages     = {15903--15935},
 year      = {2023}
}

@inproceedings{clip,
 title     = {Learning Transferable Visual Models From Natural Language Supervision},
 author    = {Alec Radford and
                  Jong Wook Kim and
                  Chris Hallacy and
                  Aditya Ramesh and
                  Gabriel Goh and
                  Sandhini Agarwal and
                  Girish Sastry and
                  Amanda Askell and
                  Pamela Mishkin and
                  Jack Clark and
                  Gretchen Krueger and
                  Ilya Sutskever},
 booktitle = {International Conference on Machine Learning},
 pages     = {8748--8763},
 year      = {2021}
}

@inproceedings{cpo,
  title={Constrained Policy Optimization},
  author={Achiam, Joshua and Held, David and Tamar, Aviv and Abbeel, Pieter},
  booktitle={International conference on machine learning},
  pages={22--31},
  year={2017}
}

@inproceedings{tdpo,
  title={Confronting Reward Overoptimization for Diffusion Models: {A} Perspective of Inductive and Primacy Biases},
  author={Ziyi Zhang and Sen Zhang and Yibing Zhan and Yong Luo and Yonggang Wen and Dacheng Tao},
  booktitle={International Conference on Machine Learning},
  pages={60396--60413},
  year={2024}
}

@inproceedings{prdp,
  title={{PRDP}: {P}roximal reward difference prediction for large-scale reward finetuning of diffusion models},
  author={Deng, Fei and Wang, Qifei and Wei, Wei and Hou, Tingbo and Grundmann, Matthias},
  booktitle={Proceedings of the IEEE/CVF Conference on Computer Vision and Pattern Recognition},
  pages={7423--7433},
  year={2024}
}

@article{rlcm,
    title={{RL} for Consistency Models: {R}eward Guided Text-to-Image Generation with Fast Inference},
    author={Oertell, Owen and Chang, Jonathan Daniel and Zhang, Yiyi and Brantley, Kiant{\'{e}} and Sun, Wen},
    journal={Reinforcement Learning Journal},
    volume={4},
    pages={1656--1673},
    year={2024}
}

@inproceedings{cm,
  title = {Consistency Models},
  author = {Song, Yang and Dhariwal, Prafulla and Chen, Mark and Sutskever, Ilya},
  booktitle = {International Conference on Machine Learning},
  pages = {32211--32252},
  year = {2023}
}

@article{lcm,
  title={{Latent Consistency Models}: {S}ynthesizing High-Resolution Images with Few-Step Inference}, 
  author={Simian Luo and Yiqin Tan and Longbo Huang and Jian Li and Hang Zhao},
  year={2023},
  journal={arXiv preprint arXiv:2310.04378},
}

@inproceedings{sdxl,
    title={{SDXL}: Improving Latent Diffusion Models for High-Resolution Image Synthesis},
    author={Dustin Podell and Zion English and Kyle Lacey and Andreas Blattmann and Tim Dockhorn and Jonas M{\"u}ller and Joe Penna and Robin Rombach},
    booktitle={International Conference on Learning Representations},
    year={2024}
}

@inproceedings{sd3,
    title={Scaling Rectified Flow Transformers for High-Resolution Image Synthesis},
    author={Patrick Esser and Sumith Kulal and Andreas Blattmann and Rahim Entezari and Jonas M{\"u}ller and Harry Saini and Yam Levi and Dominik Lorenz and Axel Sauer and Frederic Boesel and Dustin Podell and Tim Dockhorn and Zion English and Robin Rombach},
    booktitle={International Conference on Machine Learning},
    pages={12606--12633},
    year={2024}
}

@inproceedings{mvadapter,
  title={{MV-Adapter}: Multi-view Consistent Image Generation Made Easy},
  author={Huang, Zehuan and Guo, Yuan-Chen and Wang, Haoran and Yi, Ran and Ma, Lizhuang and Cao, Yan-Pei and Sheng, Lu},
  booktitle={Proceedings of the IEEE/CVF International Conference on Computer Vision},
  pages={16377--16387},
  year={2025}
}

@inproceedings{mate3d,
  title={Benchmarking and Learning Multi-Dimensional Quality Evaluator for Text-to-{3D} Generation},
  author={Yujie Zhang and Bingyang Cui and Qi Yang and Zhu Li and Yiling Xu},
  booktitle={Proceedings of the IEEE/CVF International Conference on Computer Vision},
  pages={18563--18574},
  year={2025}
}

@inproceedings{spad,
  title={{SPAD}: Spatially aware multi-view diffusers},
  author={Kant, Yash and Siarohin, Aliaksandr and Wu, Ziyi and Vasilkovsky, Michael and Qian, Guocheng and Ren, Jian and Guler, Riza Alp and Ghanem, Bernard and Tulyakov, Sergey and Gilitschenski, Igor},
  booktitle={Proceedings of the IEEE/CVF Conference on Computer Vision and Pattern Recognition},
  pages={10026--10038},
  year={2024}
}

@inproceedings{zsampling,
  title={Zigzag diffusion sampling: Diffusion models can self-improve via self-reflection},
  author={Bai, Lichen and Shao, Shitong and Zhou, Zikai and Qi, Zipeng and Xu, Zhiqiang and Xiong, Haoyi and Xie, Zeke},
  booktitle={International Conference on Learning Representations},
  year={2025}
}

@inproceedings{pso,
  title={Tuning Timestep-Distilled Diffusion Model Using Pairwise Sample Optimization},
  author={Miao, Zichen and Yang, Zhengyuan and Lin, Kevin and Wang, Ze and Liu, Zicheng and Wang, Lijuan and Qiu, Qiang},
  booktitle={Proceedings of the IEEE/CVF Conference on Computer Vision and Pattern Recognition},
  year={2025}
}

@article{sdpo,
  title={Aligning Few-Step Diffusion Models with Dense Reward Difference Learning},
  author={Zhang, Ziyi and Shen, Li and Zhang, Sen and Ye, Deheng and Luo, Yong and Shi, Miaojing and Shan, Dongjing and Du, Bo and Tao, Dacheng},
  journal={IEEE Transactions on Pattern Analysis and Machine Intelligence},
  year={2026}
}

@inproceedings{mvreward,
  title={{MVReward}: Better Aligning and Evaluating Multi-View Diffusion Models with Human Preferences},
  author={Wang, Weitao and Xu, Haoran and Yang, Yuxiao and Liu, Zhifang and Meng, Jun and Wang, Haoqian},
  booktitle={Proceedings of the AAAI Conference on Artificial Intelligence},
  volume={39},
  number={8},
  pages={7898--7906},
  year={2025}
}

@article{rebel,
  author = {Gao, Zhaolin and Chang, Jonathan D. and Zhan, Wenhao and Oertell, Owen and Swamy, Gokul and Brantley, Kiant\'{e} and Joachims, Thorsten and Bagnell, J. Andrew and Lee, Jason D. and Sun, Wen},
  journal = {Advances in Neural Information Processing Systems},
  volume = {37},
  pages = {52354--52400},
  title = {{REBEL}: Reinforcement Learning via Regressing Relative Rewards},
  year = {2024}
}

@inproceedings{dreamreward,
  title={{DreamReward}: Text-to-{3D} generation with human preference},
  author={Ye, Junliang and Liu, Fangfu and Li, Qixiu and Wang, Zhengyi and Wang, Yikai and Wang, Xinzhou and Duan, Yueqi and Zhu, Jun},
  booktitle={European Conference on Computer Vision},
  pages={259--276},
  year={2024}
}

@inproceedings{dreamalign,
  title={{DreamAlign}: Dynamic Text-to-{3D} Optimization with Human Preference Alignment},
  author={Liu, Gaofeng and Ma, Zhiyuan and Fang, Tao},
  booktitle={Proceedings of the AAAI Conference on Artificial Intelligence},
  volume={39},
  number={5},
  pages={5424--5432},
  year={2025}
}

@inproceedings{carve3d,
  title={{Carve3D}: Improving multi-view reconstruction consistency for diffusion models with rl finetuning},
  author={Xie, Desai and Li, Jiahao and Tan, Hao and Sun, Xin and Shu, Zhixin and Zhou, Yi and Bi, Sai and Pirk, S{\"o}ren and Kaufman, Arie E},
  booktitle={Proceedings of the IEEE/CVF Conference on Computer Vision and Pattern Recognition},
  pages={6369--6379},
  year={2024}
}

@inproceedings{mvdream,
  title={{MVD}ream: Multi-view Diffusion for {3D} Generation},
  author={Yichun Shi and Peng Wang and Jianglong Ye and Long Mai and Kejie Li and Xiao Yang},
  booktitle={International Conference on Learning Representations},
  year={2024}
}

@inproceedings{viewdiff,
  title={{ViewDiff}: 3d-consistent image generation with text-to-image models},
  author={H{\"o}llein, Lukas and Bo{\v{z}}i{\v{c}}, Alja{\v{z}} and M{\"u}ller, Norman and Novotny, David and Tseng, Hung-Yu and Richardt, Christian and Zollh{\"o}fer, Michael and Nie{\ss}ner, Matthias},
  booktitle={Proceedings of the IEEE/CVF Conference on Computer Vision and Pattern Recognition},
  pages={5043--5052},
  year={2024}
}

@inproceedings{instant3d,
  title={{Instant3D}: Fast Text-to-{3D} with Sparse-view Generation and Large Reconstruction Model},
  author={Jiahao Li and Hao Tan and Kai Zhang and Zexiang Xu and Fujun Luan and Yinghao Xu and Yicong Hong and Kalyan Sunkavalli and Greg Shakhnarovich and Sai Bi},
  booktitle={International Conference on Learning Representations},
  year={2024}
}

@article{zero123++,
  title={Zero123++: A single image to consistent multi-view diffusion base model},
  author={Shi, Ruoxi and Chen, Hansheng and Zhang, Zhuoyang and Liu, Minghua and Xu, Chao and Wei, Xinyue and Chen, Linghao and Zeng, Chong and Su, Hao},
  journal={arXiv preprint arXiv:2310.15110},
  year={2023}
}

@article{imagedream,
  title={{ImageDream}: Image-prompt multi-view diffusion for {3D} generation},
  author={Wang, Peng and Shi, Yichun},
  journal={arXiv preprint arXiv:2312.02201},
  year={2023}
}

@inproceedings{syncdreamer,
  title={{SyncDreamer}: Generating Multiview-consistent Images from a Single-view Image},
  author={Yuan Liu and Cheng Lin and Zijiao Zeng and Xiaoxiao Long and Lingjie Liu and Taku Komura and Wenping Wang},
  booktitle={International Conference on Learning Representations},
  year={2024}
}

@inproceedings{epidiff,
  title={{EpiDiff}: Enhancing multi-view synthesis via localized epipolar-constrained diffusion},
  author={Huang, Zehuan and Wen, Hao and Dong, Junting and Wang, Yaohui and Li, Yangguang and Chen, Xinyuan and Cao, Yan-Pei and Liang, Ding and Qiao, Yu and Dai, Bo and others},
  booktitle={Proceedings of the IEEE/CVF Conference on Computer Vision and Pattern Recognition},
  pages={9784--9794},
  year={2024}
}

@inproceedings{wonder3d,
  title={{Wonder3D}: Single image to {3D} using cross-domain diffusion},
  author={Long, Xiaoxiao and Guo, Yuan-Chen and Lin, Cheng and Liu, Yuan and Dou, Zhiyang and Liu, Lingjie and Ma, Yuexin and Zhang, Song-Hai and Habermann, Marc and Theobalt, Christian and others},
  booktitle={Proceedings of the IEEE/CVF Conference on Computer Vision and Pattern Recognition},
  pages={9970--9980},
  year={2024}
}

@inproceedings{free3d,
  title={{Free3D}: Consistent novel view synthesis without {3D} representation},
  author={Zheng, Chuanxia and Vedaldi, Andrea},
  booktitle={Proceedings of the IEEE/CVF Conference on Computer Vision and Pattern Recognition},
  pages={9720--9731},
  year={2024}
}

@inproceedings{crm,
  title={{CRM}: Single image to {3D} textured mesh with convolutional reconstruction model},
  author={Wang, Zhengyi and Wang, Yikai and Chen, Yifei and Xiang, Chendong and Chen, Shuo and Yu, Dajiang and Li, Chongxuan and Su, Hang and Zhu, Jun},
  booktitle={European Conference on Computer Vision},
  pages={57--74},
  year={2024},
}

@inproceedings{sv3d,
  title={{SV3D}: Novel multi-view synthesis and {3D} generation from a single image using latent video diffusion},
  author={Voleti, Vikram and Yao, Chun-Han and Boss, Mark and Letts, Adam and Pankratz, David and Tochilkin, Dmitry and Laforte, Christian and Rombach, Robin and Jampani, Varun},
  booktitle={European Conference on Computer Vision},
  pages={439--457},
  year={2024}
}

@inproceedings{nvsadapter,
  title={{NVS-Adapter}: Plug-and-Play Novel View Synthesis from a Single Image},
  author={Jeong, Yoonwoo and Lee, Jinwoo and Kim, Chiheon and Cho, Minsu and Lee, Doyup},
  booktitle={European Conference on Computer Vision},
  pages={449--466},
  year={2024}
}

@inproceedings{mvdiffusion++,
  title={{MVDiffusion}++: A dense high-resolution multi-view diffusion model for single or sparse-view {3D} object reconstruction},
  author={Tang, Shitao and Chen, Jiacheng and Wang, Dilin and Tang, Chengzhou and Zhang, Fuyang and Fan, Yuchen and Chandra, Vikas and Furukawa, Yasutaka and Ranjan, Rakesh},
  booktitle={European Conference on Computer Vision},
  pages={175--191},
  year={2024}
}

@article{era3d,
  title={{Era3D}: High-resolution multiview diffusion using efficient row-wise attention},
  author={Li, Peng and Liu, Yuan and Long, Xiaoxiao and Zhang, Feihu and Lin, Cheng and Li, Mengfei and Qi, Xingqun and Zhang, Shanghang and Xue, Wei and Luo, Wenhan and others},
  journal={Advances in Neural Information Processing Systems},
  volume={37},
  pages={55975--56000},
  year={2024}
}

@inproceedings{cat3d,
 title = {{CAT3D}: Create Anything in {3D} with Multi-View Diffusion Models},
 author = {Gao, Ruiqi and Ho\l y\'{n}ski, Aleksander and Henzler, Philipp and Brussee, Arthur and Martin-Brualla, Ricardo and Srinivasan, Pratul and Barron, Jonathan T. and Poole, Ben},
 booktitle = {Advances in Neural Information Processing Systems},
 pages = {75468--75494},
 volume = {37},
 year = {2024}
}

@inproceedings{
    dreamfusion,
    title={{DreamFusion}: Text-to-{3D} using {2D} Diffusion},
    author={Ben Poole and Ajay Jain and Jonathan T. Barron and Ben Mildenhall},
    booktitle={International Conference on Learning Representations},
    year={2023},
}

@inproceedings{magic3d,
  title={{Magic3D}: High-resolution text-to-{3D} content creation},
  author={Lin, Chen-Hsuan and Gao, Jun and Tang, Luming and Takikawa, Towaki and Zeng, Xiaohui and Huang, Xun and Kreis, Karsten and Fidler, Sanja and Liu, Ming-Yu and Lin, Tsung-Yi},
  booktitle={Proceedings of the IEEE/CVF Conference on Computer Vision and Pattern Recognition},
  pages={300--309},
  year={2023}
}

@inproceedings{sjc,
  title={Score jacobian chaining: Lifting pretrained {2D} diffusion models for {3D} generation},
  author={Wang, Haochen and Du, Xiaodan and Li, Jiahao and Yeh, Raymond A and Shakhnarovich, Greg},
  booktitle={Proceedings of the IEEE/CVF Conference on Computer Vision and Pattern Recognition},
  pages={12619--12629},
  year={2023}
}

@inproceedings{textmesh,
  title={{TextMesh}: Generation of realistic {3D} meshes from text prompts},
  author={Tsalicoglou, Christina and Manhardt, Fabian and Tonioni, Alessio and Niemeyer, Michael and Tombari, Federico},
  booktitle={International Conference on {3D} Vision},
  pages={1554--1563},
  year={2024}
}

@article{3dtopia,
  title={{3DTopia}: Large text-to-{3D} generation model with hybrid diffusion priors},
  author={Hong, Fangzhou and Tang, Jiaxiang and Cao, Ziang and Shi, Min and Wu, Tong and Chen, Zhaoxi and Yang, Shuai and Wang, Tengfei and Pan, Liang and Lin, Dahua and others},
  journal={arXiv preprint arXiv:2403.02234},
  year={2024}
}

@inproceedings{consistent3d,
  title={{Consistent3D}: Towards consistent high-fidelity text-to-{3D} generation with deterministic sampling prior},
  author={Wu, Zike and Zhou, Pan and Yi, Xuanyu and Yuan, Xiaoding and Zhang, Hanwang},
  booktitle={Proceedings of the IEEE/CVF Conference on Computer Vision and Pattern Recognition},
  pages={9892--9902},
  year={2024}
}

@inproceedings{latentnerf,
  title={{Latent-NeRF} for shape-guided generation of {3D} shapes and textures},
  author={Metzer, Gal and Richardson, Elad and Patashnik, Or and Giryes, Raja and Cohen-Or, Daniel},
  booktitle={Proceedings of the IEEE/CVF Conference on Computer Vision and Pattern Recognition},
  pages={12663--12673},
  year={2023}
}

@inproceedings{12345++,
  title={One-2-3-45++: Fast single image to {3D} objects with consistent multi-view generation and {3D} diffusion},
  author={Liu, Minghua and Shi, Ruoxi and Chen, Linghao and Zhang, Zhuoyang and Xu, Chao and Wei, Xinyue and Chen, Hansheng and Zeng, Chong and Gu, Jiayuan and Su, Hao},
  booktitle={Proceedings of the IEEE/CVF Conference on Computer Vision and Pattern Recognition},
  pages={10072--10083},
  year={2024}
}

@article{cdpo,
  title={Aligning Text-to-Image Diffusion Models with Constrained Reinforcement Learning},
  author={Ziyi Zhang and Sen Zhang and Li Shen and Yibing Zhan and Yong Luo and Han Hu and Bo Du and Yonggang Wen and Dacheng Tao},
  journal={IEEE Transactions on Pattern Analysis and Machine Intelligence},
  year={2025},
  volume={47},
  number={11},
  pages={9550-9562}
}
}

\clearpage
\setcounter{page}{1}
\onecolumn
\maketitlesupplementary
\appendix

\section{Detailed Discussion of Related Work}
\label{sec:related}

\textbf{Text-to-multiview generation.} MVDream~\cite{mvdream} pioneers an end-to-end text-to-multiview (T2MV) framework by adapting self-attention in pretrained T2I models to capture cross-view dependencies, enabling direct synthesis of multiview images from a text prompt. SPAD~\cite{spad} enhances this by integrating epipolar-geometry constraints into cross-view attention, while ViewDiff~\cite{viewdiff} enforces cross-view consistency through 3D volume rendering and cross-frame attention. Instead of altering the T2I backbone directly, MV-Adapter~\cite{mvadapter} injects plug-and-play adapters into the self-attention layers to facilitate end-to-end T2MV. Moreover, Instant3D~\cite{instant3d} takes an alternative end-to-end approach that compiles all views into an image grid. In contrast to these integrated methods, most other approaches~\cite{zero123++,imagedream,syncdreamer,epidiff,wonder3d,free3d,crm,sv3d,nvsadapter,mvdiffusion++,era3d,cat3d} rely on a decoupled two-stage process: first create a single reference image via a T2I model, then pass that image into a multiview diffusion model to produce novel views.

\textbf{Reinforcement learning for few-step diffusion models.} While reinforcement learning has been used to optimize few-step diffusion models, most efforts have been confined to T2I scenarios. RLCM~\cite{rlcm} adapts DDPO~\cite{ddpo}, a pioneering gradient method for diffusion models, to consistency models (CMs)~\cite{cm,lcm}. Building on this, REBEL~\cite{rebel} reframes policy optimization as regressing reward differences between trajectory pairs, with an adaptation for CMs. Moreover, Miao et al.~\cite{pso} adapt direct preference optimization to timestep-distilled diffusion models. Most recently, Zhang et al.~\cite{sdpo} introduce a dense reward difference learning objective for arbitrary few-step diffusion models beyond CMs, leveraging dense feedback for more granular policy updates. Despite their success with few-step diffusion models, these approaches offer no built-in strategy for addressing multiview challenges.

\textbf{Alignment of multiview diffusion models.} Prior works on aligning multiview diffusion models diverge from ours in task scope and objectives. MVReward~\cite{mvreward} introduces a reward-feedback learning paradigm for multiview diffusion models, but focuses on image-conditioned novel-view synthesis rather than our text-to-multiview objective. DreamReward~\cite{dreamreward} and DreamAlign~\cite{dreamalign} adapt reward-feedback and preference learning strategies, respectively, to text-to-3D pipelines built on multiview diffusion, yet both optimize the resulting 3D representation rather than the underlying multiview diffusion model. Carve3D~\cite{carve3d} employs policy gradient to finetune multiview diffusion models—much like one of our proposed baselines (i.e., MV-PG)—but relies on NeRF reconstruction-based rewards computed solely from rendered views, neglecting the evaluation of text-image alignment.

\section{Policy Optimization Paradigms for Text-to-Image Diffusion Models}
\label{appendix:t2i_rl}

As discussed in Section~\ref{sec:pre} of the main paper, several prior works have explored reinforcement learning (RL) techniques to finetune text-to-image (T2I) diffusion models using reward functions or human preference data. In this appendix, we provide detailed formulations of these policy optimization paradigms, which serve as the conceptual foundation for our multiview-aware extensions presented in Section~\ref{subsec:method1.2} of the main paper.

\subsection{MDP Formulation for T2I Diffusion}

To enable RL finetuning for text-to-image (T2I) diffusion models, prior works~\cite{sftpg,dpok,ddpo,d3po} reformulate the denoising process as a \(T\)-step MDP:
\begin{align}
    & s_t \triangleq (\mathbf{x}_{t}, \mathbf{c}), \quad a_t \triangleq \mathbf{x}_{t-1}, \quad \pi(a_t \mid s_t) \triangleq p_\theta(\mathbf{x}_{t-1} \mid \mathbf{x}_{t}, \mathbf{c}), \nonumber \\
    & P_0(s_0) = \left(p(\mathbf{c}), \mathcal{N}(0,\mathbf{I})\right), \quad P(s_{t+1} \mid s_t, a_t) = (\delta_\mathbf{c}, \delta_{a_t}), \nonumber \\
    & r(s_t, a_t) \triangleq
    \begin{cases}
        R(\mathbf{x}_0, \mathbf{c}) & \text{if}\ t = 0 \\
        0 & \text{otherwise}
    \end{cases},
    \label{eq:mdp}
\end{align}
where \(R\) is an arbitrary reward function evaluating T2I generation, \(\delta\) denotes the Dirac delta distribution, \(P_0\) and \(P\) are the initial state distribution and transition kernel, respectively. Under this MDP formulation, the objective for RL finetuning of T2I models is to learn a diffusion policy \(\pi=p_\theta\) that maximizes the expected reward over the distribution of text prompts and generated images, i.e.,
\begin{equation}
    \max_{\theta} \; \mathbb{E}_{\mathbf{c} \sim p(\mathbf{c})} \mathbb{E}_{\mathbf{x}_0 \sim p_\theta(\cdot\mid\mathbf{c})} \left[ R(\mathbf{x}_0, \mathbf{c}) \right].
    \label{eq:rl_obj}
\end{equation}

\subsection{Policy Gradient}

The above MDP enables exact computation of log-likelihoods and their gradients with respect to the diffusion model parameters. Leveraging this property, prior works on diffusion policy optimization~\cite{sftpg,dpok,ddpo} apply policy gradient (PG)~\cite{reinforce,ppo} to optimize Eq.~(\ref{eq:rl_obj}) via:
\begin{equation}
    \mathbb{E}_{p(\mathbf{c})}\mathbb{E}_{p_\theta(\mathbf{x}_{0:T} \mid \mathbf{c})}\left[ - R(\mathbf{x}_0, \mathbf{c}) \sum_{t=1}^T \nabla_\theta \log{p_{\theta} (\mathbf{x}_{t-1} \mid \mathbf{x}_t, \mathbf{c})} \right].
    \label{eq:pg_obj}
\end{equation}

\subsection{Direct Preference Optimization}

Several prior methods, such as Diffusion-DPO~\cite{wallace2024diffusion} and D3PO~\cite{d3po}, adopt direct preference optimization (DPO) to optimize diffusion models based on pairwise image preferences from a fixed human preference dataset. These methods typically update the diffusion policy by maximizing the likelihood of preferred denoising trajectories under the current policy \(p_\theta\) relative to the reference policy \(p_{\mathrm{ref}}\), leading to the following DPO objective:
\begin{equation}
    \mathbb{E}_{(\mathbf{x}_{0:T}^w, \mathbf{x}_{0:T}^l, \mathbf{c})} \left[ \log \sigma \left( \beta \sum_{t=1}^T \left( \log \frac{p_{\theta} (\mathbf{x}_{t-1}^l \mid \mathbf{x}_t^l, \mathbf{c})}{p_{\mathrm{ref}} (\mathbf{x}_{t-1}^l \mid \mathbf{x}_t^l, \mathbf{c})} - \log \frac{p_{\theta} (\mathbf{x}_{t-1}^w \mid \mathbf{x}_t^w, \mathbf{c})}{p_{\mathrm{ref}} (\mathbf{x}_{t-1}^w \mid \mathbf{x}_t^w, \mathbf{c})} \right) \right) \right],
    \label{eq:dpo_obj}
\end{equation}
where \(\mathbf{x}_{0:T}^w\) and \(\mathbf{x}_{0:T}^l\) are the preferred and rejected trajectories, \(\beta\) is a scaling factor controlling the deviation of \(p_\theta\) from \(p_{\mathrm{ref}}\), and \(\sigma\) is the sigmoid function.

\subsection{Reward Difference Learning}

Alternative approaches such as PRDP~\cite{prdp} and REBEL~\cite{rebel} reformulate policy optimization for diffusion models as a reward difference learning (RDL) problem, which penalizes the squared deviation between differences in log-likelihood ratios with corresponding differences in rewards. Specifically, RDL draws two independent denoising trajectories for the same prompt \(\mathbf{c}\), yielding state pairs \((\mathbf{x}^a_t, \mathbf{x}^b_t)\) at each timestep \(t\). The RDL objective is then defined as:
\begin{equation}
    \mathbb{E} \Bigg[ \sum_{t=1}^T \bigg( \frac{1}{\eta} \left( \log \frac{p_\theta(\mathbf{x}^a_{t-1} \mid \mathbf{x}^a_t, \mathbf{c})}{p_{\theta'}(\mathbf{x}^a_{t-1} \mid \mathbf{x}^a_t, \mathbf{c})} - \log \frac{p_\theta(\mathbf{x}^b_{t-1} \mid \mathbf{x}^b_t, \mathbf{c})}{p_{\theta'}(\mathbf{x}^b_{t-1} \mid \mathbf{x}^b_t, \mathbf{c})} \right) - \Big( R(\mathbf{x}_0^a, \mathbf{c}) - R(\mathbf{x}_0^b, \mathbf{c}) \Big) \bigg)^2 \Bigg],
    \label{eq:rdl_obj}
\end{equation}
where \(p_\theta\) and \(p_{\theta'}\) denote the current and previous policies, and \(\eta\) is a factor controlling the weight of log-likelihood ratio differences w.r.t. reward differences.

\subsection{Limitations for T2MV Tasks}

While these methods have demonstrated success in improving single-image metrics such as aesthetic quality~\cite{aes} and text-image alignment~\cite{clip,pick,hpsv2}, they are fundamentally limited when applied to few-step T2MV diffusion models. Specifically, their MDP formulation treats each view independently during policy optimization, lacking mechanisms to enforce cross-view consistency. When naively extended to T2MV by optimizing single-view rewards independently for each viewpoint, these methods tend to improve per-view image fidelity at the expense of geometric consistency across views, as their learning objectives provide no explicit cross-view supervision. This limitation motivates our development of a multiview-aware MDP formulation (Section~\ref{subsec:method1.1} of the main paper) that explicitly models the joint generation of all views and enables policy optimization via joint-view rewards that evaluate the collective quality of the multiview outputs.

\section{Additional Implementation Details}
\label{appendix:imp}

\textbf{Few-step T2MV baseline implementation.} We implement a T2MV baseline that integrates MV-Adapter~\cite{mvadapter} with the LCM-SDXL\footnote{https://huggingface.co/latent-consistency/lcm-sdxl} backbone, which is a latent consistency model~\cite{lcm} distilled from Stable Diffusion XL (SDXL)~\cite{sdxl}. In our experiments of RL finetuning for this few-step T2MV baseline, we apply LoRA~\cite{lora} to both the multiview attention layers of MV-Adapter and the U-Net layers of LCM-SDXL, enabling parameter-efficient finetuning. Each RL finetuning experiment is conducted on 4 GPUs with 24GB memory each.

\textbf{MVC-ZigAL implementation.} As outlined in Algorithm~\ref{alg:mvc_zigal}, for each training epoch, we sample a batch of $B$ text prompts and generate two sets of multiview trajectories for each prompt: one via standard sampling and the other via ZMV-Sampling with the zigzag pass applied only at the first denoising step ($t=T$). We then evaluate both single-view rewards using a T2I reward model (PickScore~\cite{pick} or HPSv2~\cite{hpsv2}) for each view and joint-view rewards using HyperScore's overall quality score~\cite{mate3d}. To stabilize training, both single-view and joint-view rewards are normalized across the entire batch to have zero mean and unit variance. Subsequently, we compute the batch-averaged joint-view reward $\bar{\mathcal{R}}_{\mathrm{mv}}$ (averaging over both standard and ZMV trajectories) and update the adaptive threshold $\tau_k$ via exponential moving average with smoothing factor $\beta_\tau = 0.99$. The Lagrange multiplier $\lambda_k$ is then updated using adaptive step sizes: $\alpha^+ = 0.1$ when the constraint is violated ($\bar{\mathcal{R}}_{\mathrm{mv}} < \tau_k$), or $\alpha^- = 0.01$ otherwise, with $\lambda_k$ capped at 5.0. Using the updated $\lambda_k$ and normalized rewards, we compute the multiview-constrained rewards that combine single-view and joint-view rewards weighted by $\lambda_k$, and then derive the multiview-constrained zigzag advantages as the differences between ZMV and standard trajectories. After storing the current model parameters as $\theta'$, we perform $N$ inner epochs where both standard and ZMV-Sampling trajectories are resampled using standard sampling with the current policy $\theta$, and the model parameters are updated by minimizing the MVC-ZigAL loss (Eq.~(\ref{eq:mvczigal_obj})) that aligns log-likelihood ratio differences with multiview-constrained zigzag advantages. For numerical stability, probabilities in log-likelihood computations are clipped to a minimum of 1e-4, and gradients are clipped by their global norm to a maximum of 5.0. Detailed hyperparameters are provided in Table~\ref{tab:hyperparams}.

\begin{table}[!ht]
    \centering
    \caption{Training hyperparameters for MVC-ZigAL. We report values for standard RL finetuning (on animal prompt set), large-scale RL finetuning (on MATE-3D prompt set), and finetuning MV-Adapter integrated with standard SDXL intead of LCM-SDXL.}
    \label{tab:hyperparams}
    \begin{tabular}{l|c|c|c}
    \toprule
    \textbf{Parameter} & \textbf{Standard Training} & \textbf{Large-Scale Training} & \textbf{SDXL Training} \\
    \midrule
    Number of views (\(V\))                     & 6          & 6          & 6          \\
    Number of inference steps (\(T\))           & 8          & 8          & 16          \\
    Image height                                & 768        & 768        & 768        \\
    Image width                                 & 768        & 768        & 768        \\
    \midrule
    Learning rate                               & 1e-4       & 5e-5       & 5e-5       \\
    Optimizer type                              & AdamW      & AdamW      & AdamW      \\
    Optimizer weight decay                      & 1e-4       & 1e-4       & 1e-4       \\
    Optimizer betas                             & (0.9, 0.999) & (0.9, 0.999) & (0.9, 0.999) \\
    Optimizer epsilon                           & 1e-8       & 1e-8       & 1e-8       \\
    \midrule
    LoRA rank                                   & 16         & 16         & 4         \\
    LoRA alpha                                  & 16         & 16         & 4         \\
    \midrule
    Random seed                                 & 42         & 42         & 42         \\
    Batches per epoch                           & 10         & 10         & 10         \\
    Number of training epochs (\(K\))                  & 70         & 100        & 70        \\
    Number of inner epochs (\(N\))                      & 1         & 1           & 1          \\
    Sampling batch size per GPU                 & 4          & 16          & 4         \\
    Training batch size per GPU                 & 1          & 1          & 1          \\
    Gradient accumulation steps                 & 8          & 16         & 16         \\
    \midrule
    Max gradient norm                           & 5.0        & 5.0        & 5.0        \\
    Clip range                                  & 1e-4       & 1e-4       & 1e-4       \\
    Log-ratio scaling factor (\(\eta\))         & 1.0        & 1.0        & 1.0        \\
    Number of zigzag passes per trajectory      & 1          & 1          & 1          \\
    Guidance scales (\(\omega_{\mathrm{high}}, \omega_{\mathrm{low}}\)) & (7.0, 1.0) & (7.0, 1.0) & (7.0, 1.0) \\
    \midrule
    Lagrange multiplier initial value (\(\lambda_0\))          & 0.0        & 0.0        & 0.0        \\
    Lagrange multiplier max value               & 5.0        & 5.0        & 5.0        \\
    Lagrange multiplier update step sizes (\(\alpha^{+}, \alpha^{-}\)) & (0.1, 0.01) & (0.1, 0.01) & (0.1, 0.01) \\
    EMA smoothing factor (\(\beta_\tau\))       & 0.99       & 0.99       & 0.99       \\
    \bottomrule
    \end{tabular}
\end{table}

\begin{algorithm}[!ht]
    \caption{MVC-ZigAL}
    \label{alg:mvc_zigal}
    \begin{algorithmic}[1]
    \STATE \textbf{Input:} T2MV diffusion model parameters \(\theta\), prompt distribution \(p(\mathbf{c})\), Lagrange multiplier initial value \(\lambda_0\), EMA smoothing factor \(\beta_\tau\), number of inference steps \(T\), number of views \(V\), batch size \(B\), number of training epochs \(K\), number of inner epochs \(N\)
    \FOR{training epoch \(k = 1, 2, \dots, K\)}
        \STATE Sample a batch of text prompts \(\{\mathbf{c}_i \sim p(\mathbf{c})\}_{i=1}^B\)
        \FOR{mini-batch \(i = 1, 2, \dots , B\)}
            \STATE Generate standard sampling trajectories across views \(\{\mathbf{x}_{0:T}^{i,s,v}\}_{v=1}^V\)
            \STATE Generate ZMV-Sampling trajectories across views \(\{\mathbf{x}_{0:T}^{i,z,v}\}_{v=1}^V\)
            \STATE Evaluate single-view rewards \(\{R(\mathbf{x}_0^{i,s,v}, \mathbf{c})\), \(R(\mathbf{x}_0^{i,z,v}, \mathbf{c})\}_{v=1}^V\)
            \STATE Evaluate joint-view rewards \(\mathcal{R}_\mathrm{mv}(\{\mathbf{x}_0^{i,s,v}\}_{v=1}^V, \mathbf{c}), \mathcal{R}_\mathrm{mv}(\{\mathbf{x}_0^{i,z,v}\}_{v=1}^V, \mathbf{c})\)
        \ENDFOR
        \STATE Normalize single-view rewards: \(R(\mathbf{x}_0^{i,*,v}, \mathbf{c}) \gets \frac{R(\mathbf{x}_0^{i,*,v}, \mathbf{c}) - \mu_R}{\sigma_R}\), \(* \in \{s, z\}\), \(\forall i, v\)
        \STATE Normalize joint-view rewards: \(\mathcal{R}_\mathrm{mv}(\{\mathbf{x}_0^{i,*,v}\}_{v=1}^V, \mathbf{c}) \gets \frac{\mathcal{R}_\mathrm{mv}(\{\mathbf{x}_0^{i,*,v}\}_{v=1}^V, \mathbf{c}) - \mu_\mathrm{mv}}{\sigma_\mathrm{mv}}\), \(* \in \{s, z\}\), \(\forall i\)
        \STATE Compute batch-averaged joint-view reward \( \bar{\mathcal{R}}_\mathrm{mv} = \frac{1}{2B} \sum_{i=1}^B \left( \mathcal{R}_\mathrm{mv}(\{\mathbf{x}_0^{i,s,v}\}_{v=1}^V, \mathbf{c}) + \mathcal{R}_\mathrm{mv}(\{\mathbf{x}_0^{i,z,v}\}_{v=1}^V, \mathbf{c}) \right) \)
        \IF{\(k == 1\)}
            \STATE Initialize \(\tau_{k}\): \(\tau_{k} \gets \bar{\mathcal{R}}_\mathrm{mv}\)
        \ELSE
            \STATE Update \(\tau_{k}\): \(\tau_{k} \gets \beta_\tau \tau_{k-1} + (1 - \beta_\tau) \bar{\mathcal{R}}_\mathrm{mv}\)
        \ENDIF
        \IF{\(\bar{\mathcal{R}}_\mathrm{mv} < \tau_{k}\)}
            \STATE \(\lambda_{k} \gets \max(\lambda_{k-1} + \alpha^+ (\tau_{k-1} - \bar{\mathcal{R}}_\mathrm{mv}), 0)\)
        \ELSE
            \STATE \(\lambda_{k} \gets \max(\lambda_{k-1} + \alpha^- (\tau_{k-1} - \bar{\mathcal{R}}_\mathrm{mv}), 0)\)
        \ENDIF
        \FOR{mini-batch \(i = 1, 2, \dots , B\)}
            \STATE Compute multiview-constrained rewards \(\{\mathcal{R}_\mathrm{mvc}(\mathbf{x}_0^{i,s,v}, \mathbf{c}), \mathcal{R}_\mathrm{mvc}(\mathbf{x}_0^{i,z,v}, \mathbf{c})\}_{v=1}^V\) via:
            \begin{equation*}
                \mathcal{R}_\mathrm{mvc}(\mathbf{x}_0^{i,*,v}, \mathbf{c}) = \frac{R(\mathbf{x}_0^{i,*,v}, \mathbf{c}) + \lambda_{k} \mathcal{R}_\mathrm{mv}(\{\mathbf{x}_0^{i,*,v}\}_{v=1}^V, \mathbf{c})}{1 + \lambda_{k}}, \quad * \in \{s, z\}
            \end{equation*}
            \STATE Compute multiview-constrained zigzag advantages \(\{\mathcal{A}_\mathrm{mvc}(\mathbf{x}_0^{i,z,v}, \mathbf{x}_0^{i,s,v},\mathbf{c})\}_{v=1}^V\) via:
            \begin{equation*}
                \mathcal{A}_\mathrm{mvc}(\mathbf{x}_0^{i,z,v}, \mathbf{x}_0^{i,s,v},\mathbf{c}) = \mathcal{R}_\mathrm{mvc}(\mathbf{x}_0^{i,z,v}, \mathbf{c}) - \mathcal{R}_\mathrm{mvc}(\mathbf{x}_0^{i,s,v}, \mathbf{c})
            \end{equation*}
        \ENDFOR
        \STATE Store current model parameters: \(\theta' \gets \theta\)
        \FOR{inner epoch \(n = 1, 2, \dots, N\)}
            \STATE Re-sample both standard and ZMV-Sampling trajectories using standard sampling with \(\theta\)
            \STATE Compute log-ratio differences and the MVC-ZigAL loss via Eq.~(\ref{eq:mvczigal_obj})
            \STATE Update \(\theta\) by minimizing the MVC-ZigAL loss
        \ENDFOR
    \ENDFOR
    \STATE \textbf{Output:} Optimized T2MV diffusion model parameters \(\theta\) 
    \end{algorithmic}
\end{algorithm}

\section{Approximate Inversion Operation for ZMV-Sampling}
\label{appendix:inversion}

In the context of our zigzag multiview sampling (ZMV-Sampling), the approximate inversion operation is a critical component of the three-step zigzag pass designed to enhance both cross-view and text-image alignment in T2MV diffusion models. This operation, denoted as \( q_\theta\left(\cdot \mid \{\mathbf{x}_{t-1}^v, \mathbf{e}_v\}_{v=1}^V, \mathbf{c}; \omega_{\mathrm{low}}\right) \), generates a noisier sample \( \{\tilde{\mathbf{x}}_t^v\}_{v=1}^V \) at timestep \( t \) from the partially denoised latent \( \{\mathbf{x}_{t-1}^v\}_{v=1}^V \), using a low guidance scale \( \omega_{\mathrm{low}} \) (typically 1.0) that minimizes conditional influence.

\textbf{Motivation.} The approximate inversion operation leverages the guidance gap between denoising and inversion processes. The alternation between low-guidance inversion and high-guidance re-denoising creates a contrastive mechanism: weakly aligned features are suppressed during inversion, while strongly aligned features persist and are amplified during re-denoising. This iterative refinement effectively filters out condition-irrelevant noise and strengthens condition-consistent representations, ultimately improving both text-image alignment and cross-view consistency.

\textbf{Technical formulation.} Inspired by DDIM inversion~\cite{ddim}, which deterministically reverses the denoising process to map a less noisy latent \( \mathbf{x}_{t-1} \) to a noisier latent \( \mathbf{x}_t \), we define our approximate inversion operation for the T2MV setting as follows:
\begin{equation}
    \{\tilde{\mathbf{x}}_t^v\}_{v=1}^V \sim q_\theta\left(\cdot \mid \{\mathbf{x}_{t-1}^v, \mathbf{e}_v\}_{v=1}^V, \mathbf{c}; \omega_{\mathrm{low}}\right),
\end{equation}
where \( q_\theta \) is parameterized by the same diffusion model used for denoising, but employs a low guidance scale \( \omega_{\mathrm{low}} \). To implement this, we first predict the noise component at timestep \( t \) via:
\begin{equation}
    \epsilon_\theta^t\left(\{\mathbf{x}_{t-1}^v\}_{v=1}^V\right) = \epsilon_\theta\left(\{\mathbf{x}_{t-1}^v, \mathbf{e}_v\}_{v=1}^V, \mathbf{c}, t; \omega_{\mathrm{low}}\right),
\end{equation}
and then compute the noisier latent via:
\begin{equation}
    \{\tilde{\mathbf{x}}_t^v\}_{v=1}^V = \sqrt{\alpha_t} f_\theta\left(\{\mathbf{x}_{t-1}^v, \mathbf{e}_v\}_{v=1}^V, \mathbf{c}, t\right) + \sqrt{1 - \alpha_t} \epsilon_\theta^t\left(\{\mathbf{x}_{t-1}^v\}_{v=1}^V\right),
\end{equation}
where \( f_\theta \) is the predicted denoised sample function, and \( \alpha_t \) is the noise schedule coefficient at timestep \(t\). Following zigzag diffusion~\cite{zsampling}, we approximate \( \epsilon_\theta^t\left(\{\mathbf{x}_{t-1}^v\}_{v=1}^V\right) \approx \epsilon_\theta^t\left(\{\tilde{\mathbf{x}}_{t}^v\}_{v=1}^V\right) \) for simplicity. While this may introduce errors when \( \mathbf{x}_{t-1}^v \) and \( \tilde{\mathbf{x}}_t^v \) differ substantially, it provides a plug-and-play solution compatible with diverse diffusion architectures, including few-step variants like latent consistency models.

\begin{table*}[!ht]
    \caption{\textbf{Out-of-distribution evaluation on the MATE-3D benchmark.} The number of function evaluations (NFE) indicates the total number of diffusion model passes required per inference. Our MVC-ZigAL consistently outperforms the SOTA T2MV diffusion baselines (SPAD~\cite{spad} and MV-Adapter~\cite{mvadapter}) and all SOTA methods listed in the MATE-3D benchmark across all metrics at low NFEs. Results of SOTA methods are directly taken from the MATE-3D benchmark.}
    \label{tab:mate3d}
    \centering
    \resizebox{\linewidth}{!}{
    \begin{tabular}{lcccccccc}
    \toprule
    \textbf{Method} & \textbf{NFE} & \multicolumn{4}{c}{\textbf{HyperScore}} & \textbf{ImageReward} & \textbf{HPSv2} & \textbf{PickScore} \\
    & & Alignment & Geometry & Texture & Overall \\
    \midrule
    \rowcolor{gray!15}
    \textit{(a) Results of MATE-3D SOTA methods} & & & & & & & & \\
    \midrule
    SJC~\cite{sjc} & 50 & 4.02 & 2.96 & 2.65 & 3.01 & - & - & - \\
    Latent-NeRF~\cite{latentnerf} & 50 & 4.53 & 3.44 & 3.43 & 3.56 & - & - & - \\
    3DTopia~\cite{3dtopia} & 200 & 4.36 & 4.04 & 4.37 & 4.05 & - & - & - \\
    TextMesh~\cite{textmesh} & - & 4.47 & 4.25 & 4.10 & 4.12 & - & - & - \\
    DreamFusion~\cite{dreamfusion} & 50 & 4.60 & 4.16 & 4.02 & 4.13 & - & - & - \\
    Consistent3D~\cite{consistent3d} & - & 4.84 & 4.06 & 4.01 & 4.08 & - & - & - \\
    Magic3D~\cite{magic3d} & 50 & 5.46 & 4.93 & 4.75 & 4.85 & - & - & - \\
    One-2-3-45++~\cite{12345++} & 50 & \underline{7.06} & 6.68 & 6.65 & 6.71 & - & - & - \\
    \midrule
    \rowcolor{gray!15}
    \textit{(b) Results of T2MV diffusion baselines} & & & & & & & & \\
    \midrule
    SPAD~\cite{spad} & 100 & 6.61 & 6.57 & 6.32 & 6.43 & -0.628 & 0.256 & 0.201 \\
    MV-Adapter~\cite{mvadapter} (\textbf{Baseline}) & \textbf{4} & 6.47 & 6.84 & 6.38 & 6.48 & -0.876 & 0.250 & 0.204 \\
    & \underline{8} & 6.69 & 6.97 & 6.54 & 6.67 & -0.846 & 0.252 & 0.204 \\
    \midrule
    \rowcolor{blue!10}
    \textit{(c) Results of standard RL finetuning} & & & & & & & & \\
    \midrule
    MVC-ZigAL (\textbf{Ours}) & \textbf{4} & 6.86 & 7.34 & 6.76 & 6.92 & 0.158 & 0.266 & 0.216 \\
    & \underline{8} & 7.01 & 7.39 & 6.89 & 7.04 &  0.180 & 0.268 & 0.217 \\
    \midrule
    \rowcolor{red!10}
    \textit{(d) Results of large-scale RL finetuning} & & & & & & & & \\
    \midrule
    MVC-ZigAL (\textbf{Ours}) & \textbf{4} & 7.00 & \underline{7.61} & \underline{6.97} & \underline{7.14} & \underline{0.836} & \underline{0.277} & \underline{0.223} \\
    & \underline{8} & \textbf{7.13} & \textbf{7.66} & \textbf{7.10} & \textbf{7.24} & \textbf{0.865} & \textbf{0.280} & \textbf{0.224} \\
    \bottomrule
    \end{tabular}
    }
\end{table*}

\section{Out-of-Distribution Evaluations on MATE-3D Benchmark}
\label{appendix:mate3d}

To further validate the effectiveness of our MVC-ZigAL method, we conduct out-of-distribution evaluations on the MATE-3D benchmark~\cite{mate3d}, which contains 160 diverse and unseen text prompts. As shown in Table~\ref{tab:mate3d}, our MVC-ZigAL finetuned model consistently outperforms the SOTA T2MV diffusion baselines (SPAD~\cite{spad} and MV-Adapter~\cite{mvadapter}) and all SOTA methods listed in the MATE-3D benchmark, while operating at significantly lower number of function evaluations (NFE = 4 or 8). Notably, our finetuned model achieve substantial improvements not only in training rewards (HyperScore's overall quality, PickScore, and HPSv2) but also in out-of-domain metrics (HyperScore's alignment, geometry, and texture quality, as well as ImageReward). These results demonstrate the strong generalization capability of our MVC-ZigAL in enhancing the generative performance of few-step T2MV diffusion models across diverse and unseen text prompts.

\section{Extension to Large-Scale RL Finetuning}
\label{appendix:large}

To further assess the scalability of our MVC-ZigAL framework, we conduct large-scale RL finetuning experiments using a significantly expanded prompt set. Specifically, we finetune the MV-Adapter (LCM-SDXL) baseline using our MVC-ZigAL framework on the MATE-3D prompt set~\cite{mate3d}, which contains 160 diverse prompts spanning various object categories. As presented in Table~\ref{tab:mate3d}, scaling up the training data yields significantly higher scores across all evaluated metrics, demonstrating that our MVC-ZigAL framework effectively scales with larger and more diverse training data to further enhance the generative capabilities of few-step T2MV diffusion models. Notably, the large-scale finetuned model achieves a remarkable reversal in ImageReward from -0.846 (baseline) to 0.865, representing a dramatic +1.711 improvement that reflects substantially enhanced overall image quality and text-prompt alignment. This quantitative gain is accompanied by a clear qualitative improvement, as illustrated in Figure~\ref{fig:vis_mate3d} to~\ref{fig:vis_6views_56}. The hyperparameters used for large-scale finetuning are provided in Table~\ref{tab:hyperparams}.

\section{Extension to Hyper-Step T2MV Diffusion Models}
\label{appendix:sdxl}

\begin{figure*}[!ht]
    \centering
    \includegraphics[width=0.23\textwidth]{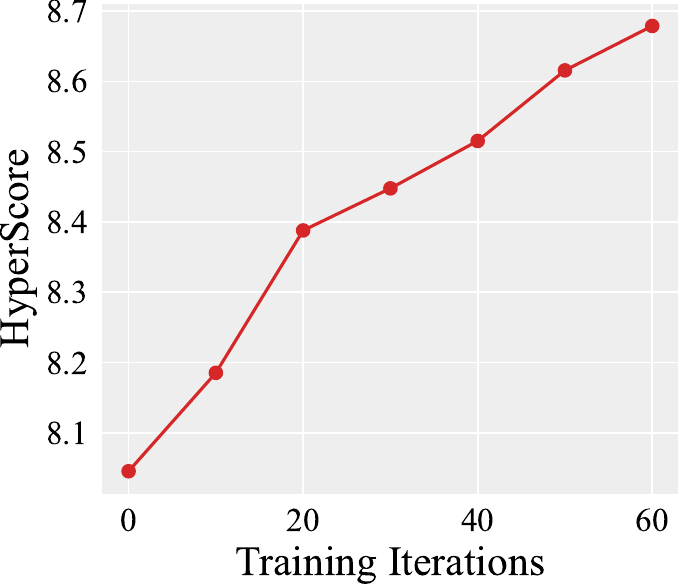}
    \hskip 0.2cm
    \includegraphics[width=0.23\textwidth]{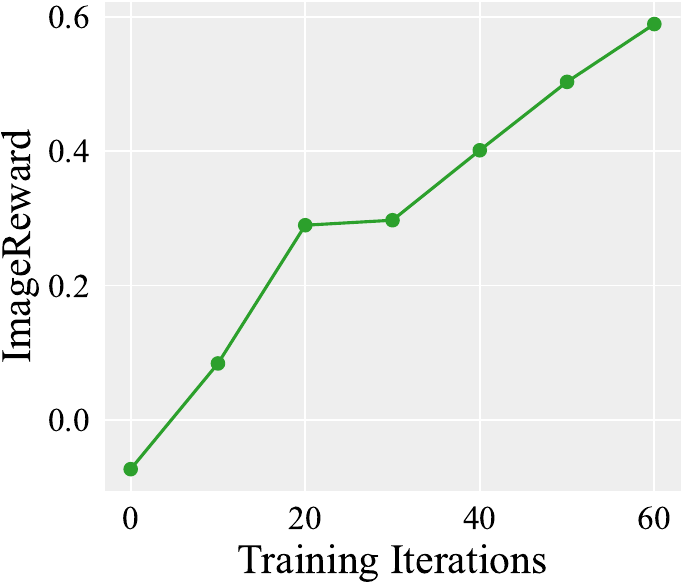}
    \hskip 0.2cm
    \includegraphics[width=0.24\textwidth]{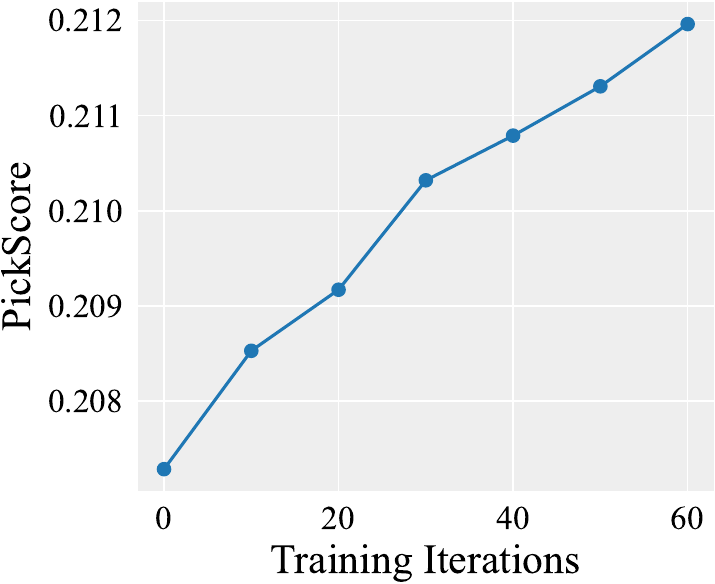}
    \hskip 0.2cm
    \includegraphics[width=0.24\textwidth]{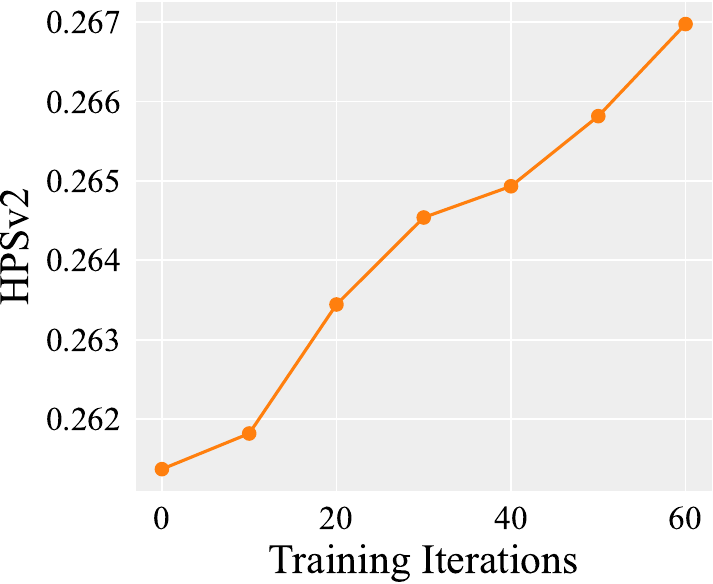}
    \caption{\textbf{Reward curves of our MVC-ZigAL} during RL finetuning of MV-Adapter (\textbf{SDXL}).}
    \label{fig:sdxl}
\end{figure*}

While our main experiments focus on finetuning few-step T2MV diffusion models, our proposed MVC-ZigAL framework is also applicable to hyper-step T2MV diffusion models, such as those based on Stable Diffusion XL (SDXL)~\cite{sdxl}. To demonstrate this, we conduct additional RL finetuning experiments using MV-Adapter integrated with SDXL as the backbone diffusion model. As shown in Figure~\ref{fig:sdxl}, our MVC-ZigAL consistently improves all reward metrics (HyperScore, ImageReward, PickScore, and HPSv2) during the RL finetuning process. These results confirm the versatility and effectiveness of MVC-ZigAL in enhancing the generative ability of T2MV diffusion models, regardless of the underlying diffusion architecture or number of inference steps. The hyperparameters used for finetuning the SDXL-based model are provided in Table~\ref{tab:hyperparams}.

\section{Additional Qualitative Comparisons}
\label{appendix:vis}

\begin{figure}[!ht]
    \centering
    \includegraphics[width=\textwidth]{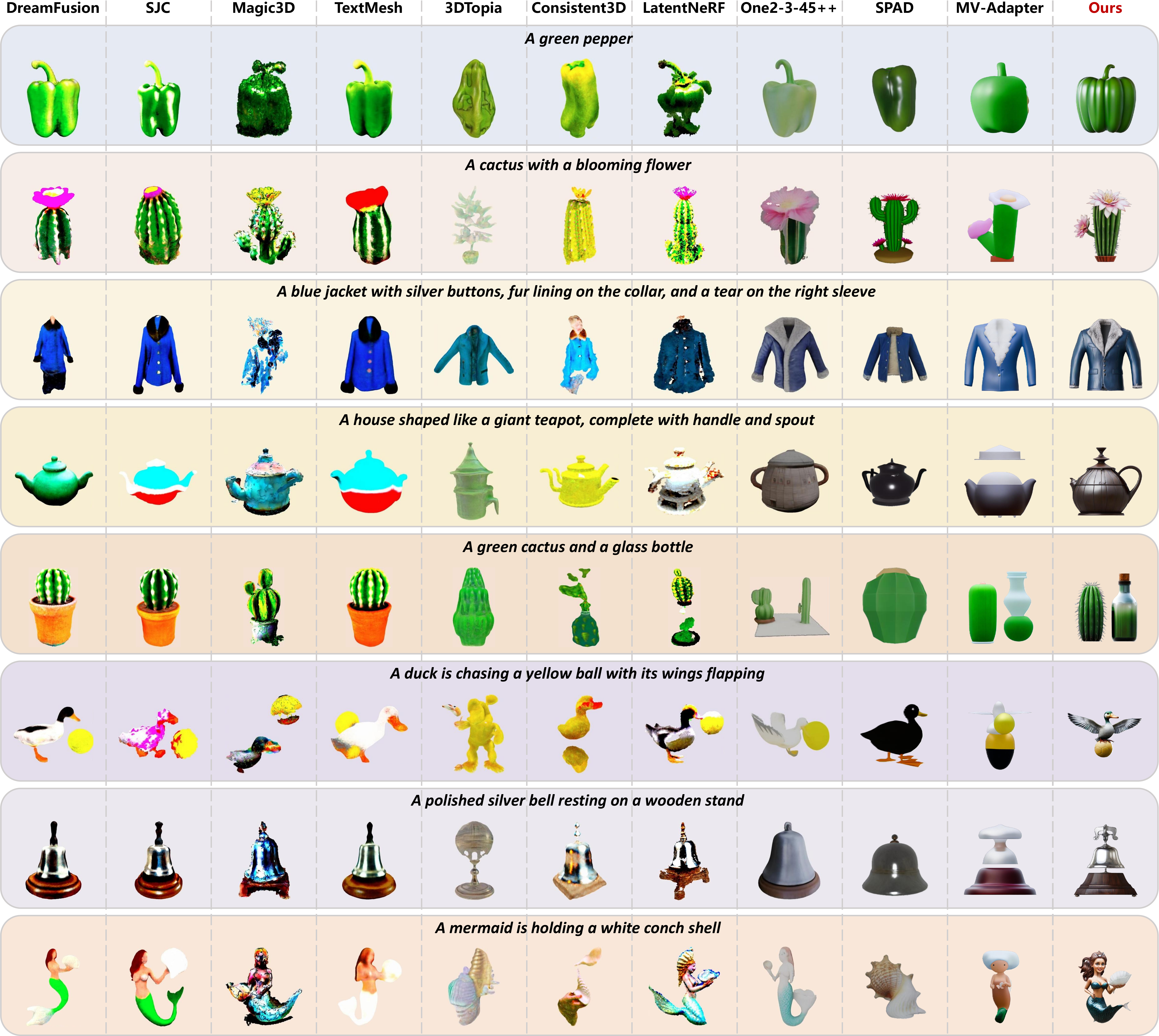}
    \caption{\textbf{Additional qualitative comparison with SOTA methods on the MATE-3D benchmark.} Results of both MV-Adapter and our MVC-ZigAL are generated using \textbf{8 inference steps (NFE = 8)}. Results of SPAD~\cite{spad} are generated using 100 inference steps (NFE = 100). Results of other methods are directly taken from the MATE-3D benchmark. Our MVC-ZigAL produces images with significantly enhanced visual quality, better text-prompt alignment, and more realistic textures, while requiring much fewer inference steps than SPAD and other SOTA methods. These qualitative comparisons further demonstrate the capability of our MVC-ZigAL finetuned model in achieving state-of-the-art T2MV generative performance.}
    \label{fig:vis_mate3d}
\end{figure}

\begin{figure}[!ht]
    \centering
    \includegraphics[width=0.97\textwidth]{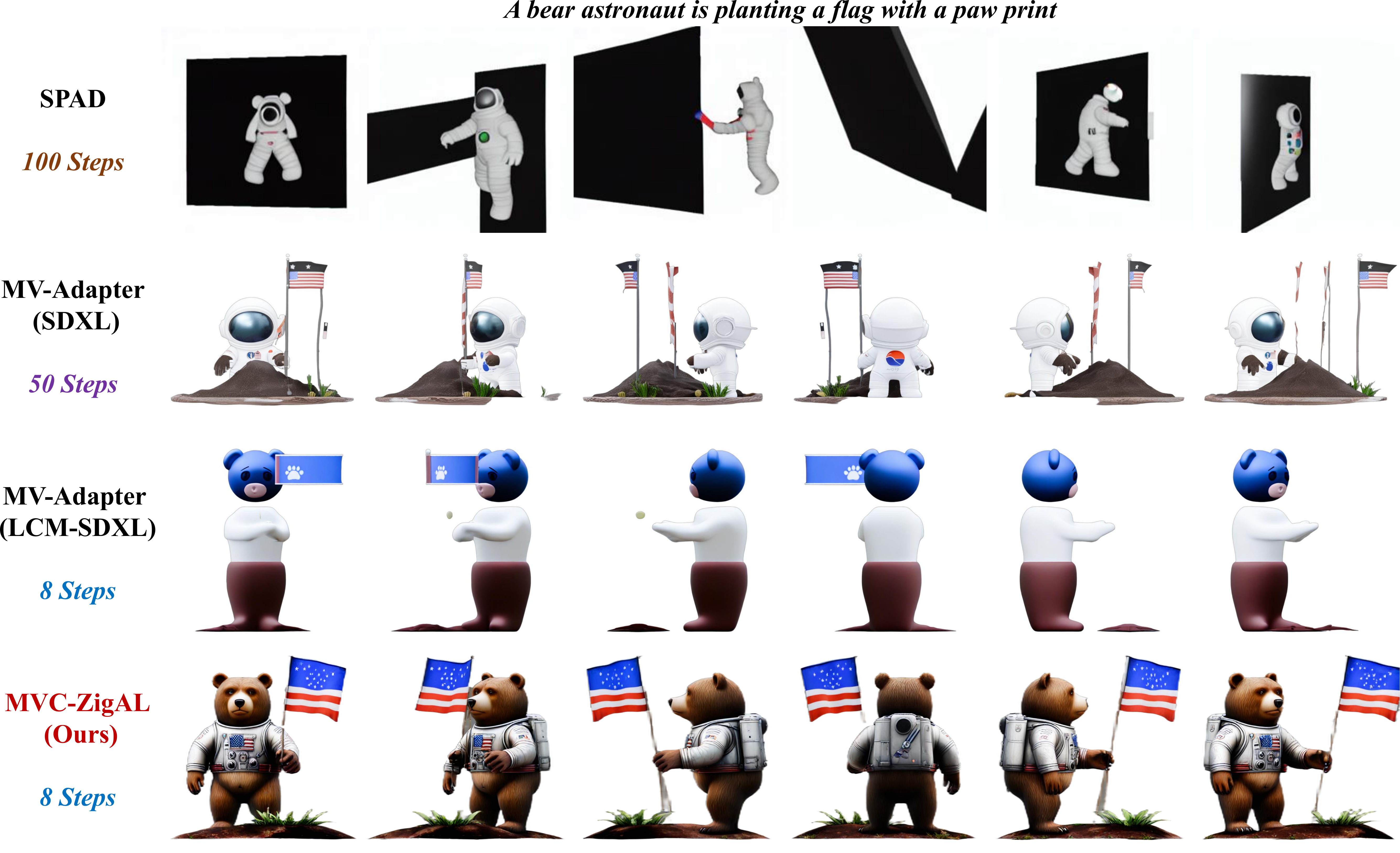}
     \\[20pt]
    \includegraphics[width=0.97\textwidth]{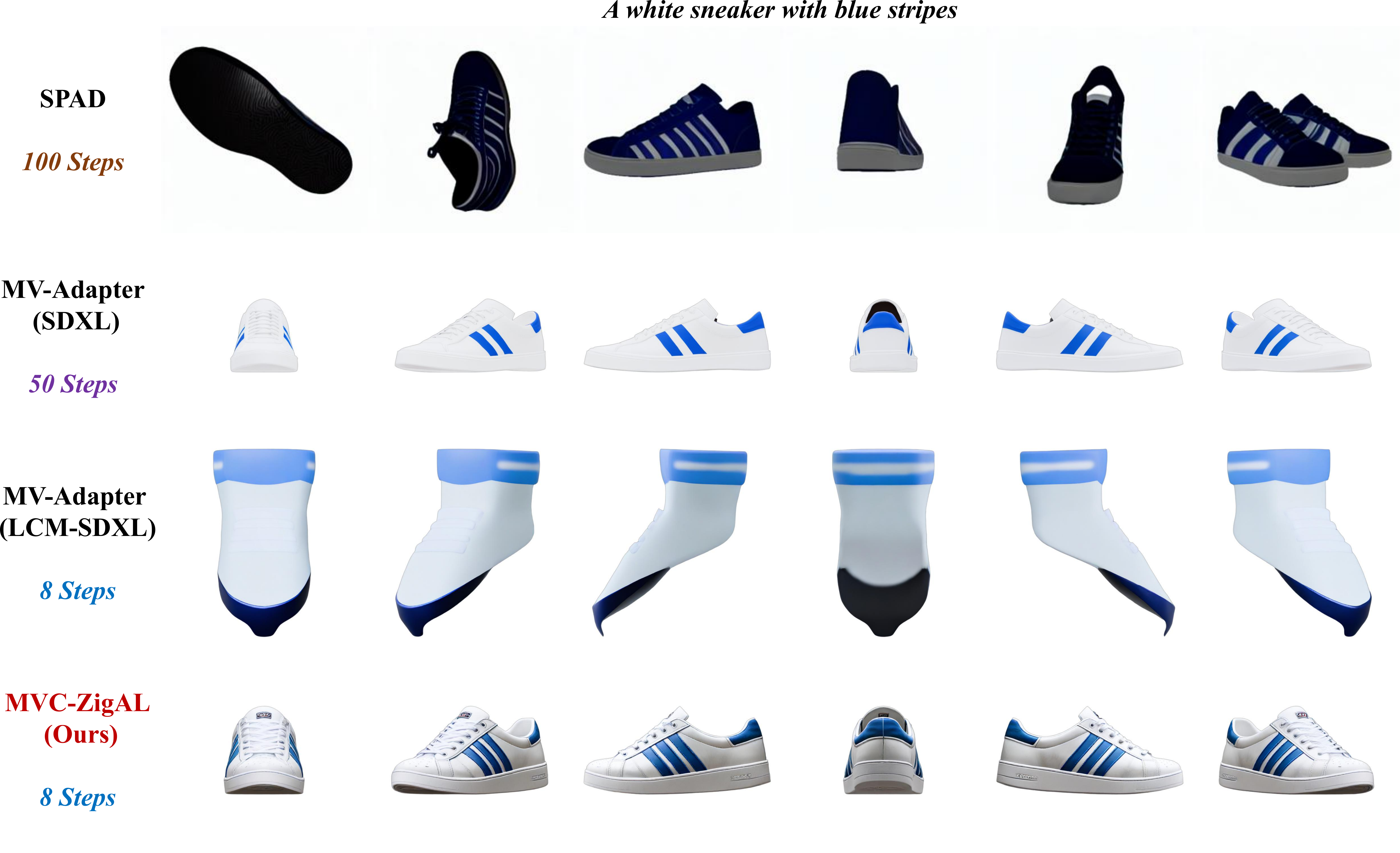}
    \caption{\textbf{Additional qualitative comparison on MATE-3D prompts.} Results from SPAD, the MV-Adapter baselines, and our MVC-ZigAL finetuned model are shown across all generated 6 views.}
    \label{fig:vis_6views_12}
\end{figure}

\begin{figure}[!ht]
    \centering
    \includegraphics[width=0.97\textwidth]{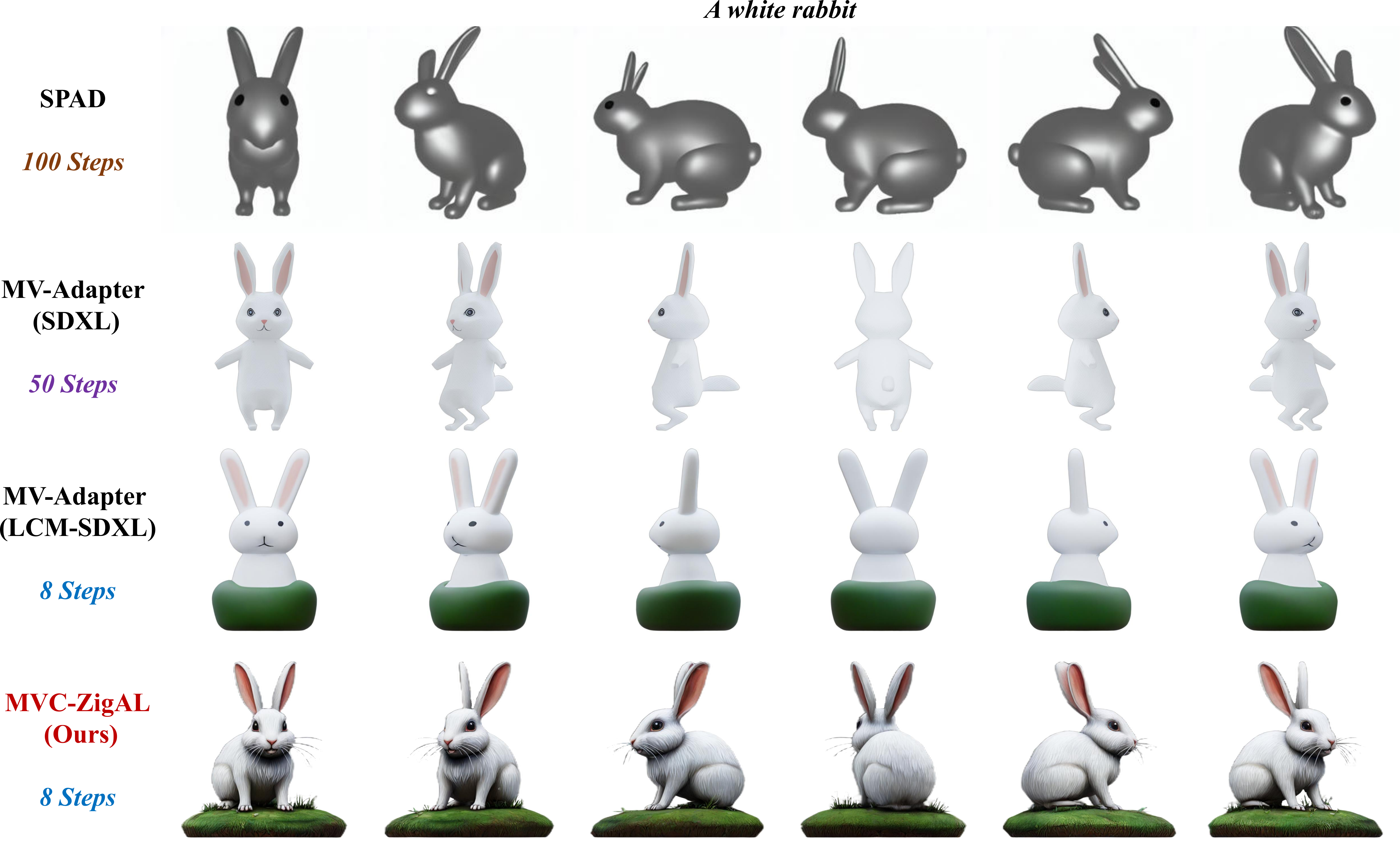}
     \\[20pt]
    \includegraphics[width=0.97\textwidth]{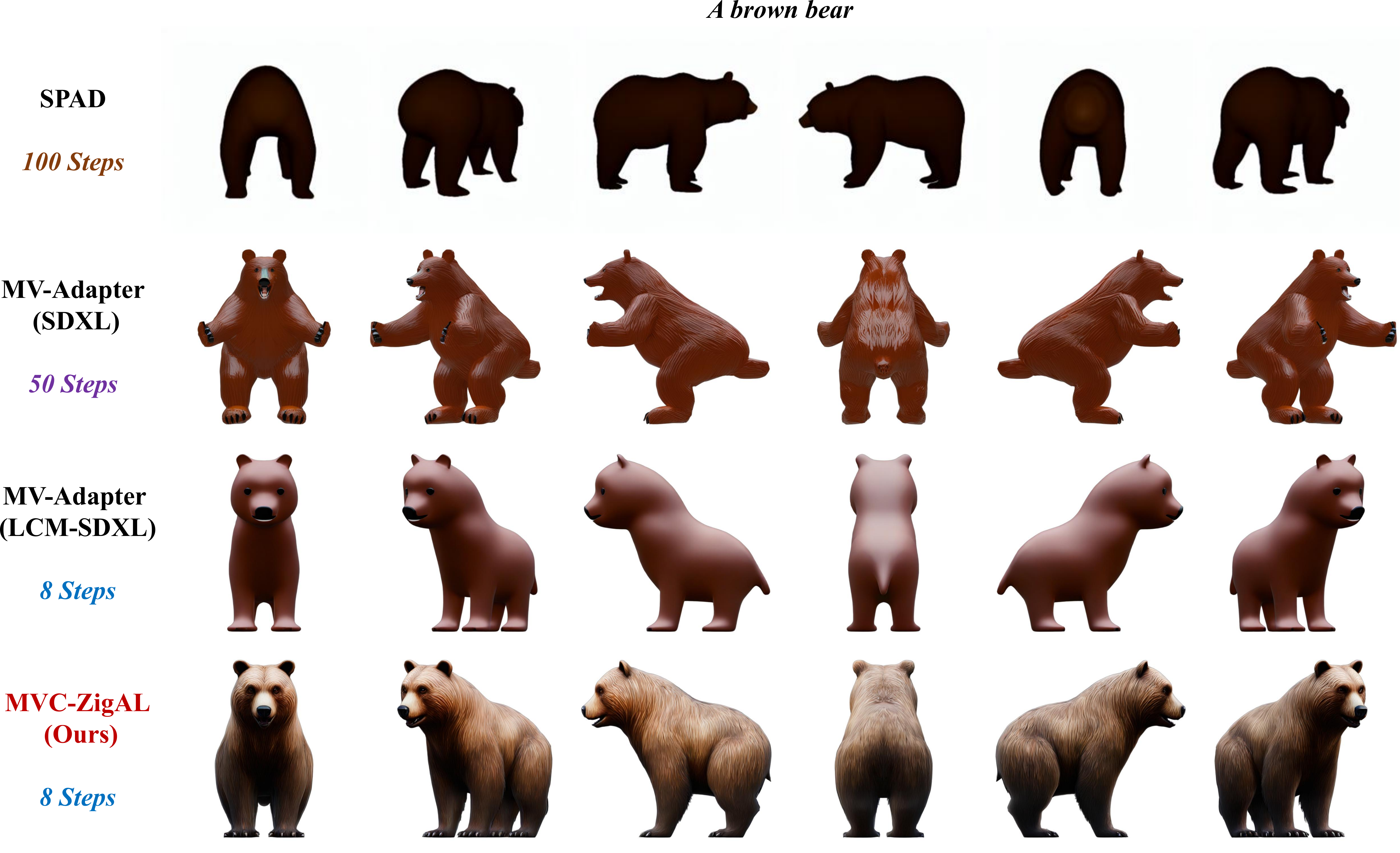}
    \caption{\textbf{Additional qualitative comparison on MATE-3D prompts.} Results from SPAD, the MV-Adapter baselines, and our MVC-ZigAL finetuned model are shown across all generated 6 views.}
    \label{fig:vis_6views_34}
\end{figure}

\begin{figure}[!ht]
    \centering
    \includegraphics[width=0.97\textwidth]{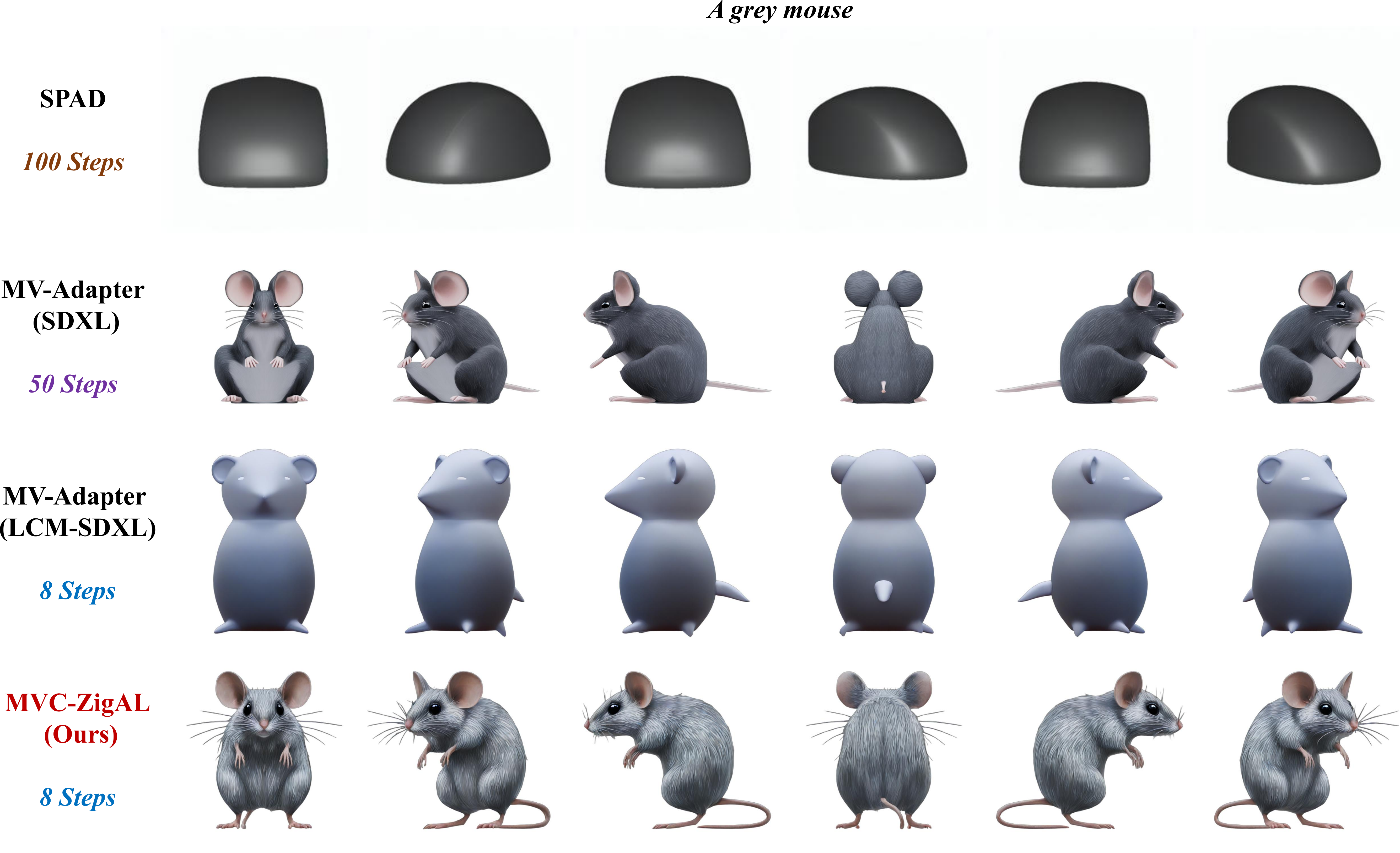}
     \\[20pt]
    \includegraphics[width=0.97\textwidth]{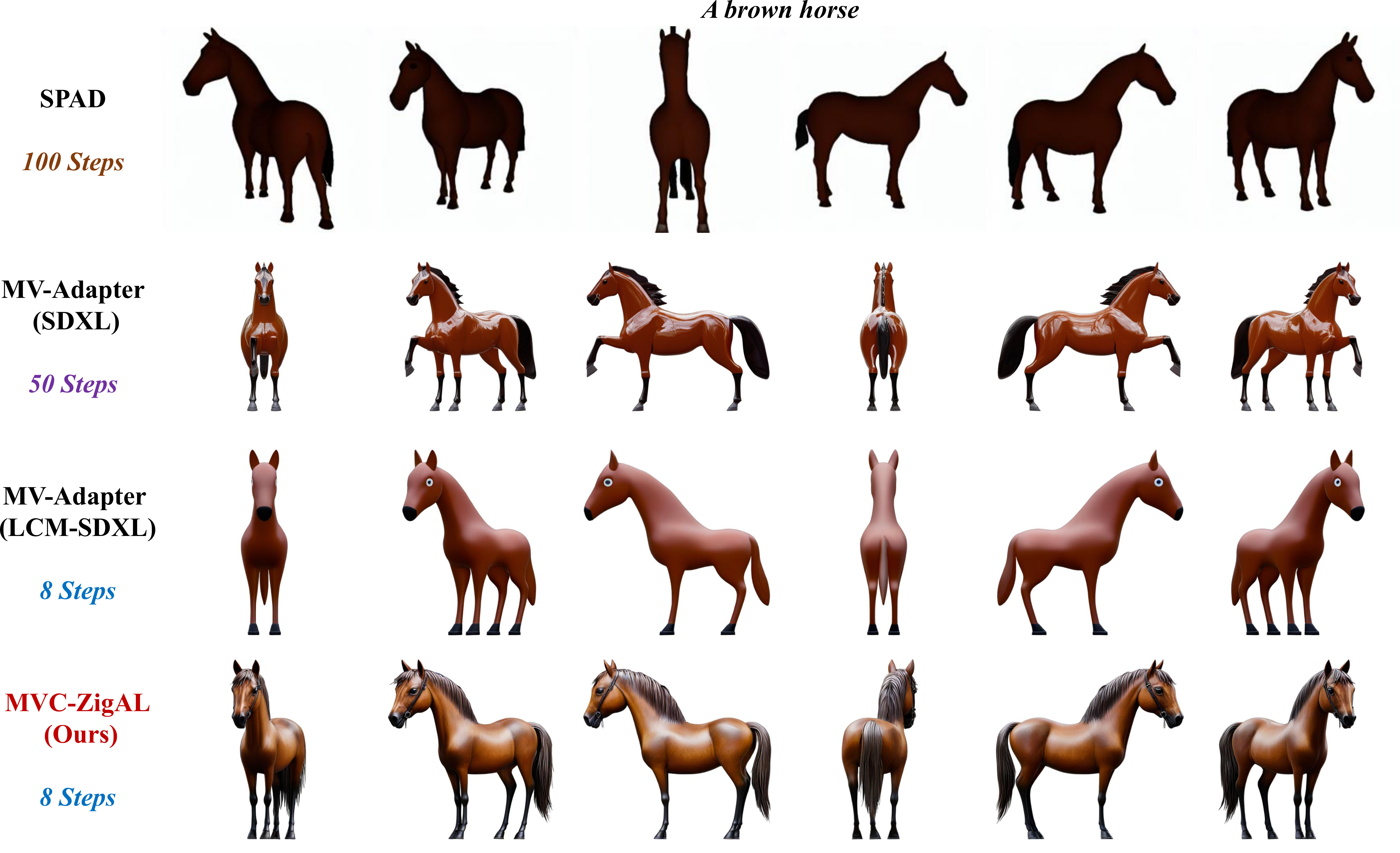}
    \caption{\textbf{Additional qualitative comparison on MATE-3D prompts.} Results from SPAD, the MV-Adapter baselines, and our MVC-ZigAL finetuned model are shown across all generated 6 views.}
    \label{fig:vis_6views_56}
\end{figure}

Figure~\ref{fig:vis_mate3d} presents additional qualitative comparisons between our MVC-ZigAL finetuned model and various SOTA methods listed in the MATE-3D benchmark~\cite{mate3d}. Due to space constraints, Figure~\ref{fig:vis_mate3d} only visualizes one view per prompt. To provide a more comprehensive comparison, Figures~\ref{fig:vis_6views_12} to~\ref{fig:vis_6views_56} showcase the complete set of 6 generated views for selected MATE-3D prompts, comparing results from SPAD~\cite{spad}, the MV-Adapter baselines (using either SDXL or LCM-SDXL as backbone), and our MVC-ZigAL finetuned model. These visualizations clearly demonstrate the superior T2MV generative quality achieved by our approach compared to existing SOTA methods.

\begin{figure}[!ht]
    \centering
    \includegraphics[width=0.8\textwidth]{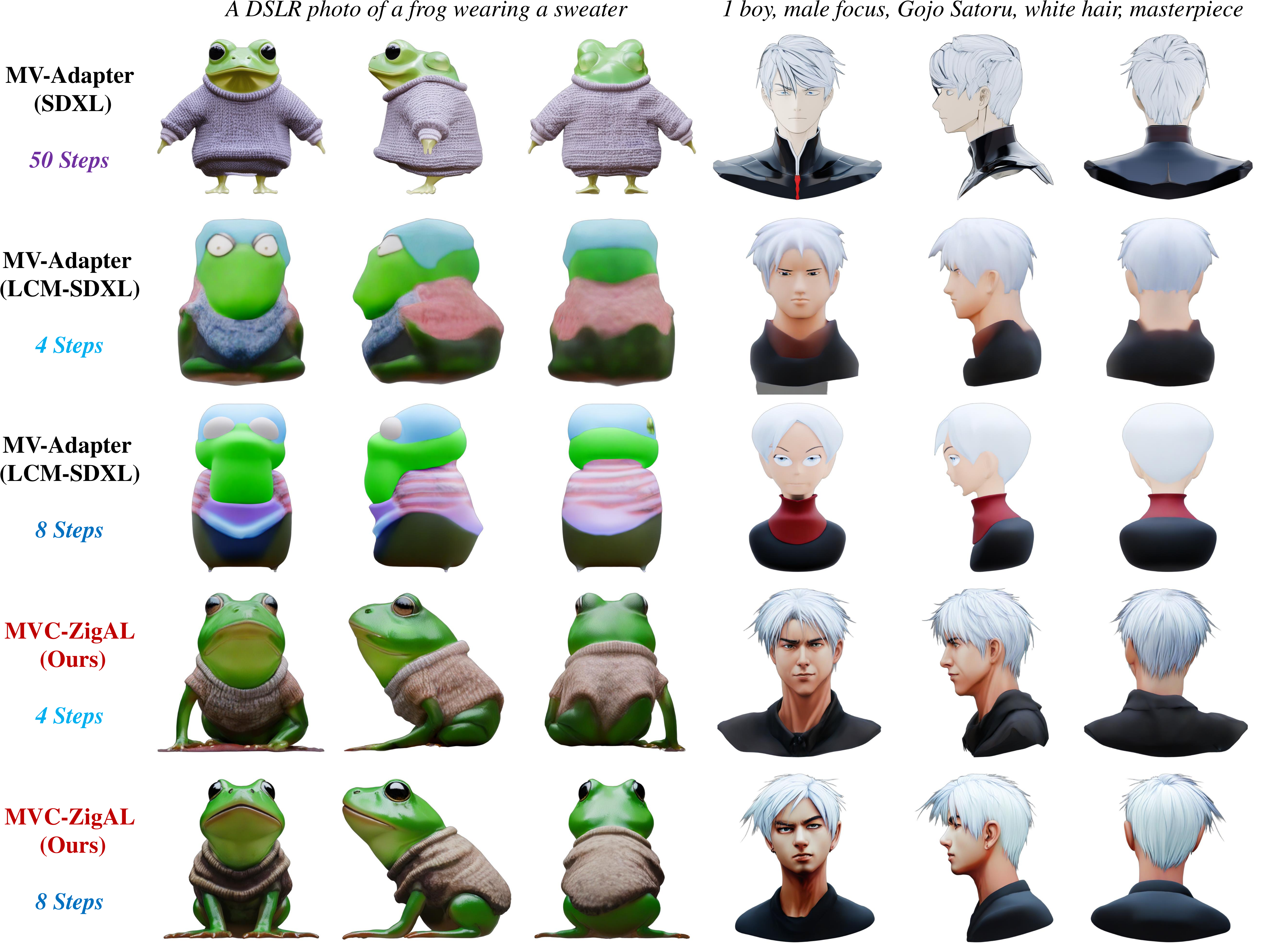}
     \\[10pt]
    \includegraphics[width=0.8\textwidth]{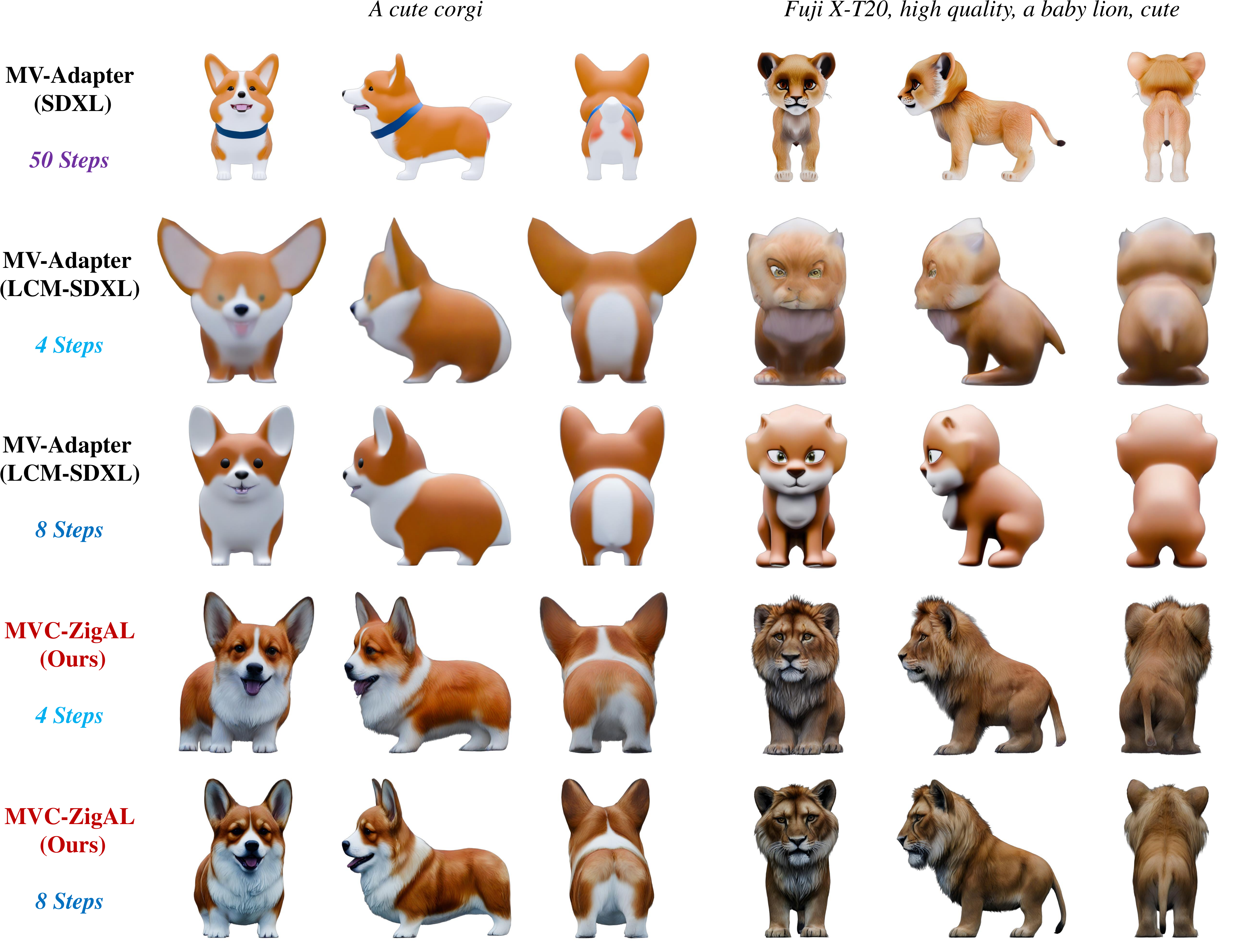}
    \caption{\textbf{Additional qualitative comparison on unseen prompts} between MV-Adapter and our MVC-ZigAL finetuned model.}
    \label{fig:vis_app1}
\end{figure}

\begin{figure}[!ht]
    \centering
    \includegraphics[width=0.8\textwidth]{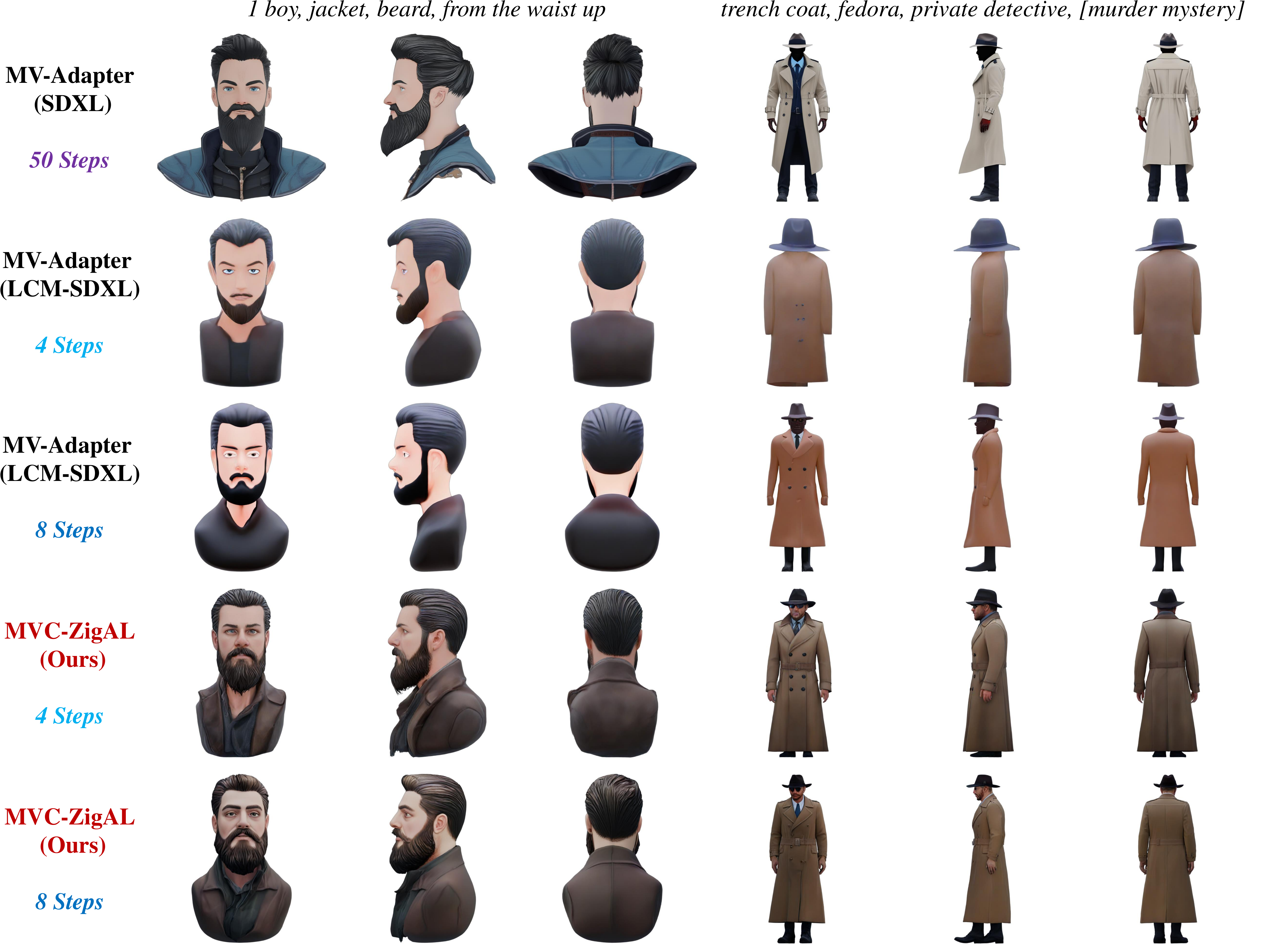}
     \\[10pt]
    \includegraphics[width=0.8\textwidth]{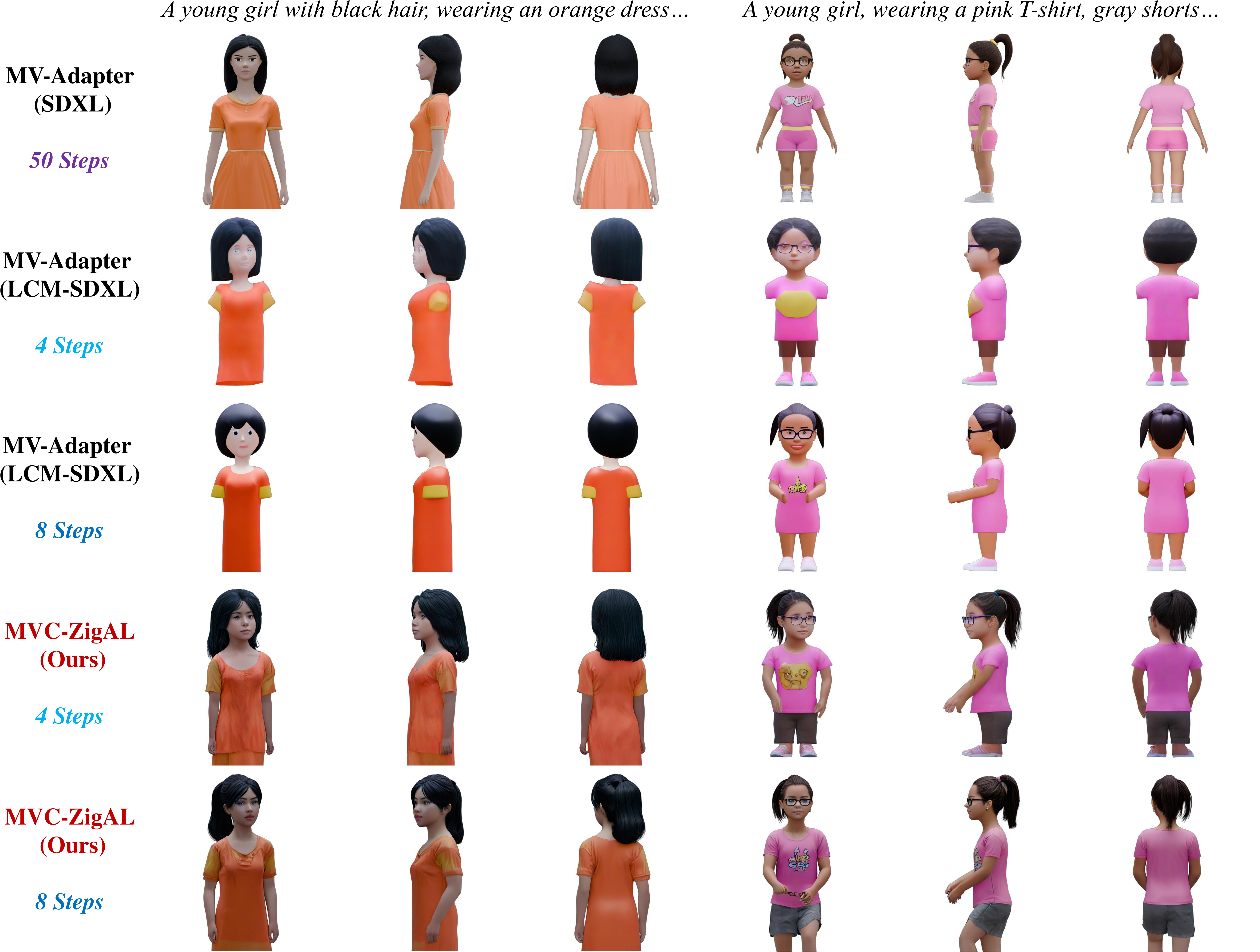}
    \caption{\textbf{Additional qualitative comparison on unseen prompts} between MV-Adapter and our MVC-ZigAL finetuned model.}
    \label{fig:vis_app2}
\end{figure}

\begin{figure}[!ht]
    \centering
    \includegraphics[width=0.8\textwidth]{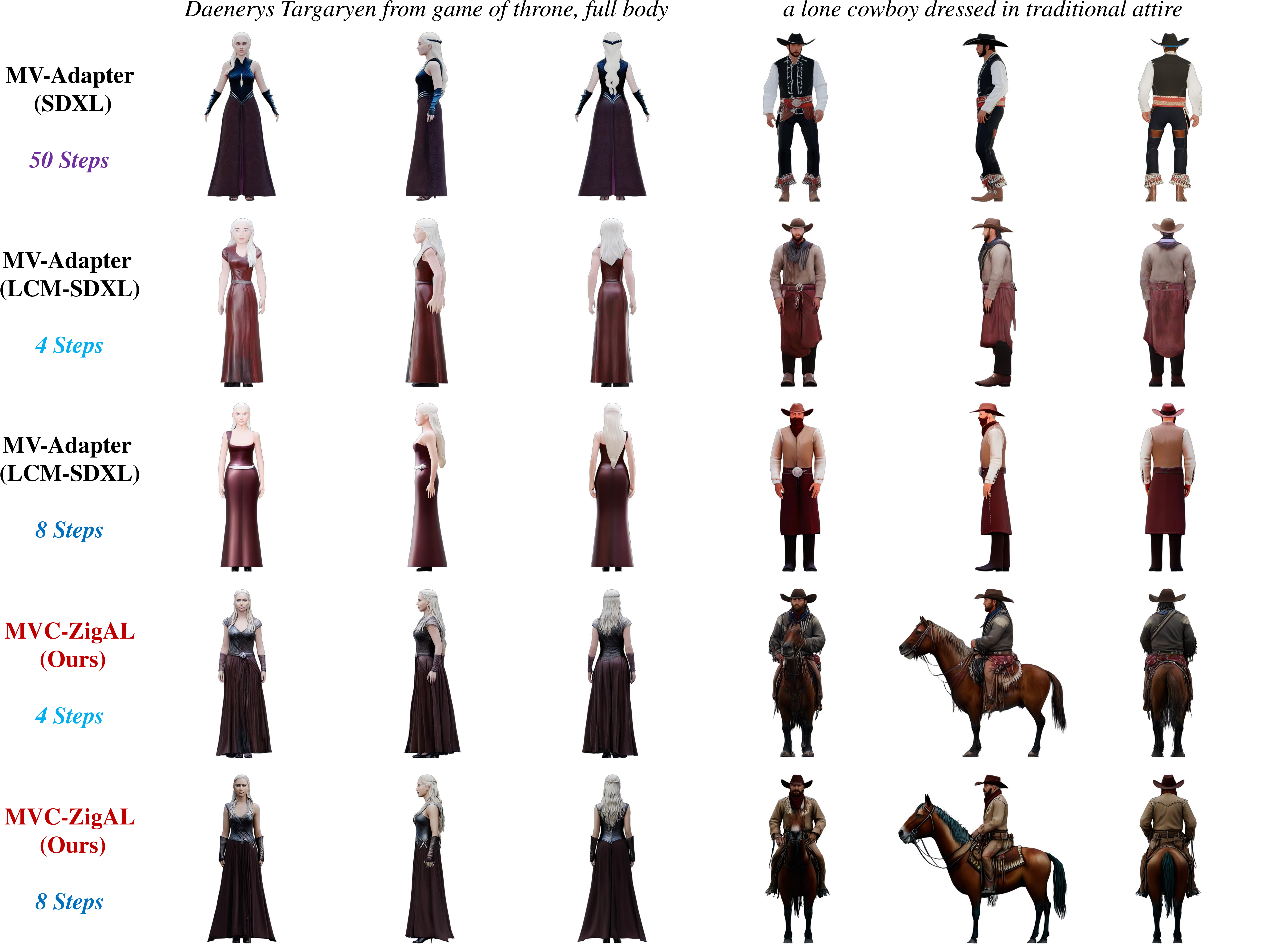}
     \\[10pt]
    \includegraphics[width=0.8\textwidth]{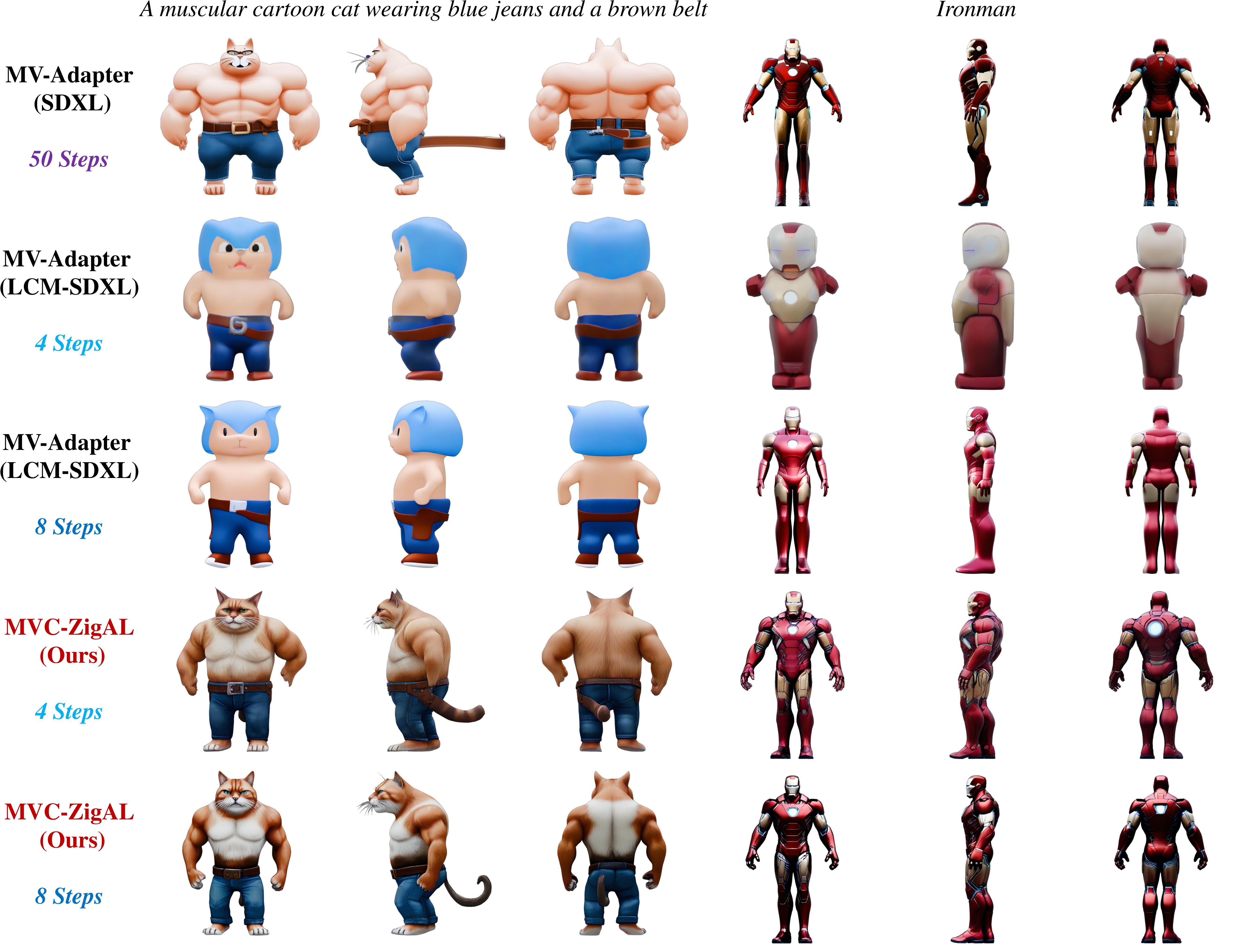}
    \caption{\textbf{Additional qualitative comparison on unseen prompts} between MV-Adapter and our MVC-ZigAL finetuned model.}
    \label{fig:vis_app3}
\end{figure}

\begin{figure}[!ht]
    \centering
    \includegraphics[width=0.8\textwidth]{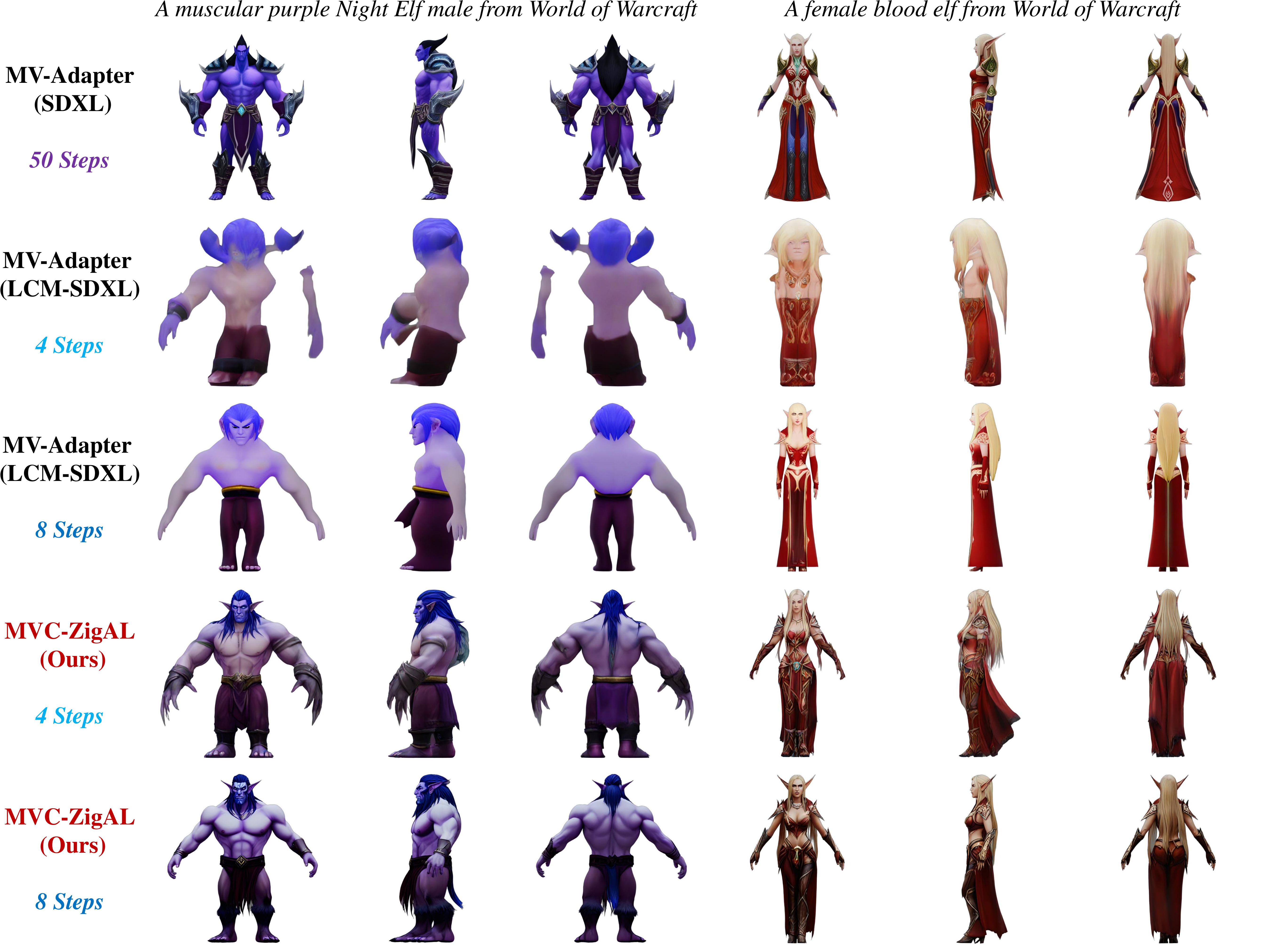}
     \\[10pt]
    \includegraphics[width=0.8\textwidth]{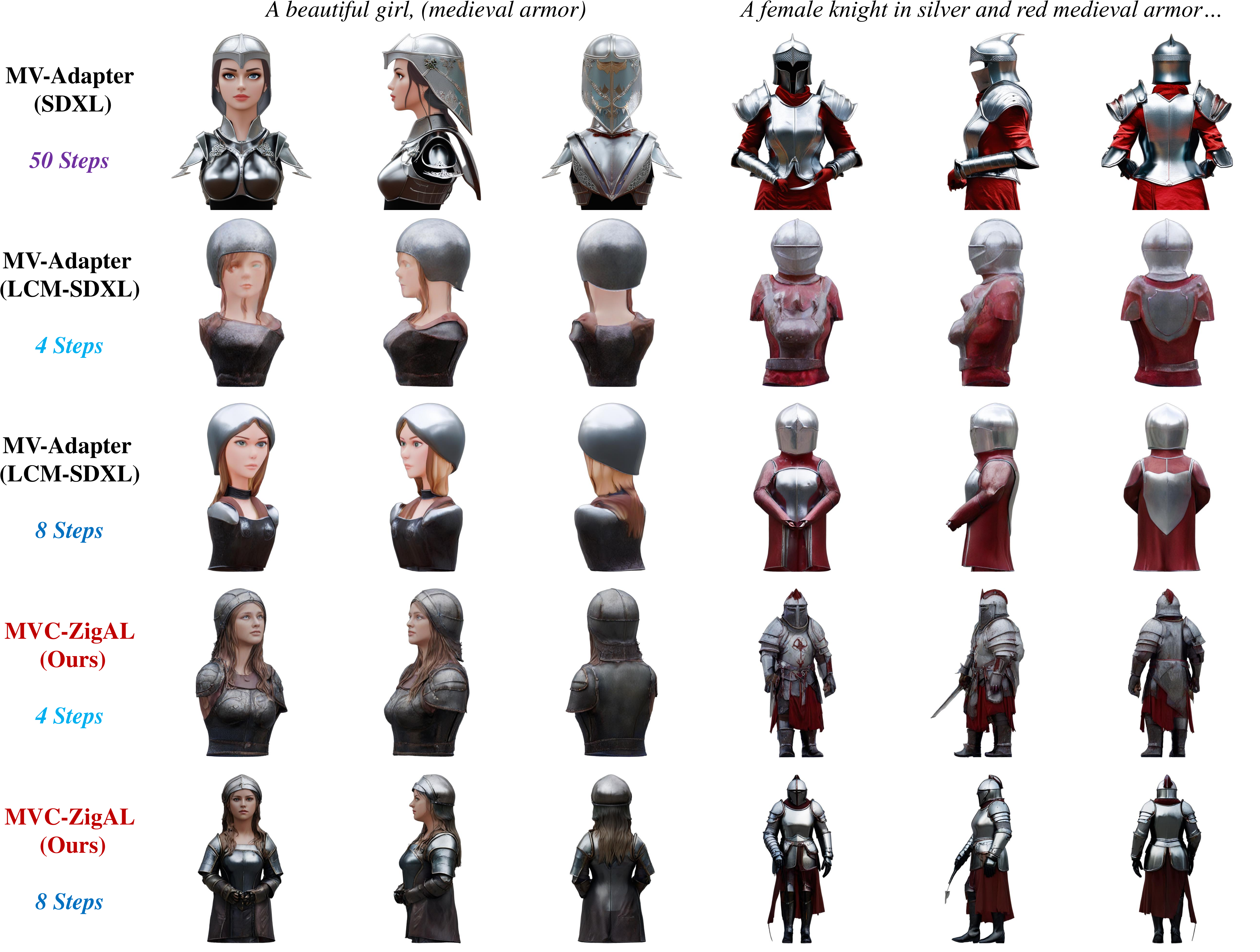}
    \caption{\textbf{Additional qualitative comparison on unseen prompts} between MV-Adapter and our MVC-ZigAL finetuned model.}
    \label{fig:vis_app4}
\end{figure}

\begin{figure}[!ht]
    \centering
    \includegraphics[width=0.8\textwidth]{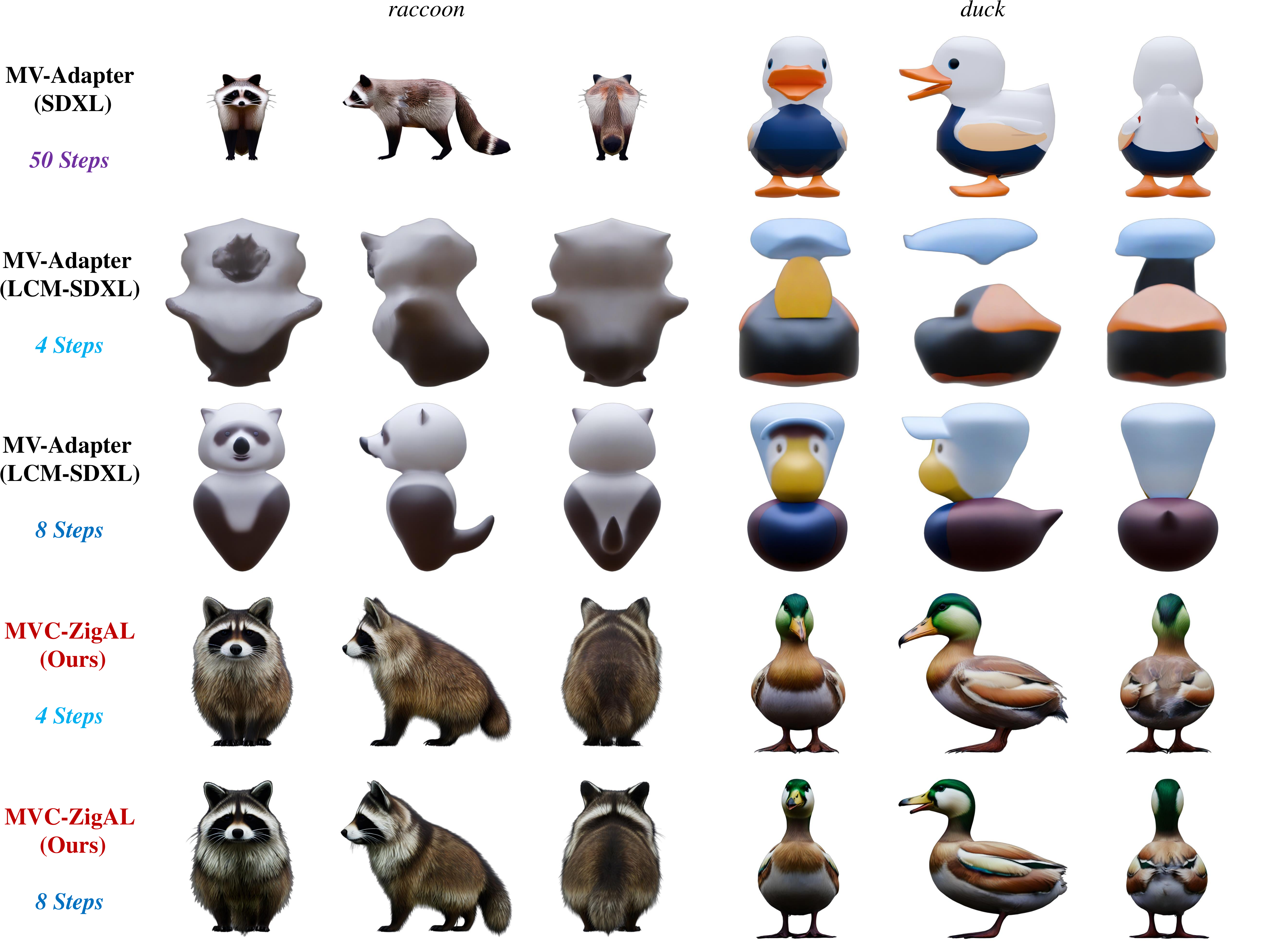}
     \\[10pt]
    \includegraphics[width=0.8\textwidth]{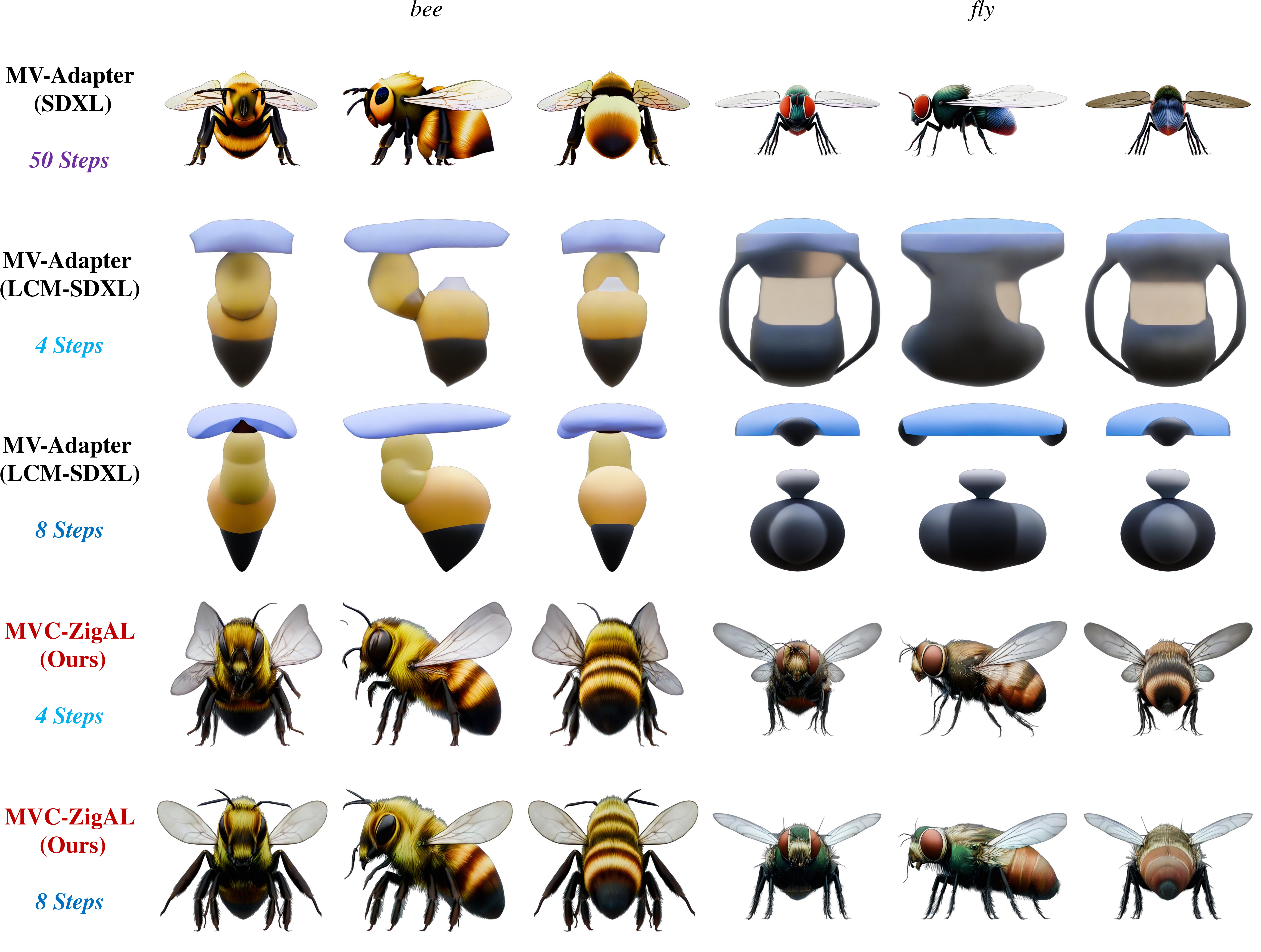}
    \caption{\textbf{Additional qualitative comparison on training prompts} between MV-Adapter and our MVC-ZigAL finetuned model.}
    \label{fig:vis_app5}
\end{figure}

Moreover, Figures~\ref{fig:vis_app1} to~\ref{fig:vis_app5} extend the evaluation to more diverse test scenarios, including both unseen prompts (not included in either the animal prompt set or the MATE-3D prompt set) and training prompts, under a more challenging 4-step setting (NFE = 4). To ensure fair comparison, all results from both the MV-Adapter baselines and our MVC-ZigAL finetuned model are generated using the same random seed (42). These additional qualitative results further demonstrate the effectiveness and robustness of our MVC-ZigAL framework in enhancing the generative performance of few-step T2MV diffusion models across diverse text prompts and inference settings.

\newpage

\newpage

\newpage

\newpage

\end{document}